\documentclass[12pt]{article}
\usepackage{hyperref}   

\usepackage{amsmath, amssymb, amsthm}
\usepackage{graphicx}
\usepackage{eufrak}
\newcommand{\ra}{\rightarrow}

\newcommand{\Lra}{\Leftrightarrow}
\newcommand{\bC}{{\bf C}}
\newcommand{\PP}{{\bf PP}}
\newcommand{\PPi}{{\bf PP^{\sim}}}
\newcommand{\NTPP}{{\bf NTPP}}
\newcommand{\NTPPi}{{\bf NTPP^{\sim}}}
\newcommand{\TPP}{{\bf TPP}}
\newcommand{\TPPi}{{\bf TPP^{\sim}}}
\newcommand{\bO}{{\bf O}}
\newcommand{\bH}{{\bf H}}
\newcommand{\bM}{{\bf M}}
\newcommand{\bN}{{\bf N}}
\newcommand{\bR}{{\bf R}}
\newcommand{\bS}{{\bf S}}
\newcommand{\PON}{{\bf PON}}
\newcommand{\POD}{{\bf POD}}

\newcommand{\PODY}{{\bf PODY}}
\newcommand{\PODZ}{{\bf PODZ}}

\newcommand{\EC}{{\bf EC}}
\newcommand{\ECN}{{\bf ECN}}
\newcommand{\ECNB}{{\bf ECNB}}
\newcommand{\ECD}{{\bf ECD}}
\newcommand{\DC}{{\bf DC}}

\newcommand{\DN}{{\bf DN}}

\newcommand{\bomega}{$\mathfrak{B}_\omega$}
\newcommand{\RC}{\mathrm{RC}}

\begin{document}
\newtheorem{lemma}{Lemma}[section]
\newtheorem{prop}{Proposition}[section]
\newtheorem{thm}{Theorem}[section]
\newtheorem{coro}{Corollary}[section]
\newtheorem{ex}{Example}[section]
\newtheorem{defi}{Definition}[section]
\newtheorem{remark}{Remark}[section]
\baselineskip 0.25in

\title{\Large\bf Relational reasoning in the Region Connection
Calculus
\thanks{This work is an unpublished paper which was
revised on September 10, 2003. The early web site
\url{http://www.compscipreprints.com/comp/Preprint/sanjiang/20030910/1}
is now inaccessible.}}
\author{{Y{\sc ongming} LI$^a$,
\thanks{Email:liyongm@snnu.edu.cn}
S{\sc anjiang} LI$^b$,
\thanks{Email: lisanjiang@tsinghua.edu.cn}
 M{\sc ingsheng} YING}$^b$
 \thanks{Email: yingmsh@tsinghua.edu.cn}
 \\
  {\small $^a$Department of Mathematics,
  Shaanxi Normal University, Xi'an, 710062,China;}\\
  {\small $^b$State Key Laboratory of Intelligent Technology and
  Systems,}\\
{\small Department of Computer Science and Technology,}\\
{\small Tsinghua University, Beijing 100084, China.}}
\date{}
\maketitle
\begin{center}
\begin{minipage}{140mm}
\centerline{\bf Abstract}\vskip 3mm {This paper is mainly
concerned with the relation-algebraical aspects of the well-known
Region Connection Calculus (RCC). We show that the contact
relation algebra (CRA) of certain RCC model is not atomic complete
and hence infinite. So in general an extensional composition table
for the RCC cannot be obtained by simply refining the RCC8
relations. After having shown that each RCC model is a consistent
model of the RCC11 {\bf CT}, we give an exhaustive investigation
about extensional interpretation of the RCC11 {\bf CT}, where we
attach a superscript $^\times$ to a cell entry in the table if and
only if extensional interpretation is impossible for this entry.
More important, we show the complemented closed disk algebra is a
representation for the relation algebra determined by the RCC11
table. The domain of this algebra contains two classes of regions,
the closed disks and closures of their complements in the real
plane, and the contact relation is standard Whiteheadean contact
(i.e. $a\bC b$ iff $a\cap b\neq\emptyset$).}
\end{minipage}
\end{center}

\vskip 3mm
\section{Introduction}
Since the mid-1970's the relational methods has become a
fundamental conceptual and methodological tool in computer
science. The wide-ranging diversity and applicability of
relational methods has been demonstrated by series of RelMiCS
seminars (\emph{International Seminar on Relational Methods in
Computer Science}). Relation algebra has been used as a basis for
analyzing, modeling or resolving several computer science problems
such as program specification, heuristic approaches for program
derivation, automatic prover design, database and software
decomposition, program fault tolerance, testing, data abstraction
and information coding, and last but not least qualitative spatial
reasoning. For a detailed overview we invite the reader to consult
\cite{BKS97,JG99,OS01,DFJM01}.

Qualitative spatial reasoning (QSR) is an important subfield of
{\bf AI} which is concerned with the qualitative aspects of
representing and reasoning about spatial entities. A large part of
contemporary qualitative spatial reasoning is based on the
behavior of ``part of " and ``connection" (or ``contact")
relations in various domains \cite{EH91,CBGG97}, and the
expressive power, consistency and complexity of relational
reasoning has become an important object of study in QSR.

Rather than give to attention to all the various systems existing
on the market, we shall focus on one of the most widely referenced
formalism for QSR, the {\it Region Connection Calculus} (RCC). RCC
was initially described by Randell, Cohn and Cui in
\cite{RC89,RCuCo92}, which is intended to provide a logical
framework for incorporating spatial reasoning into {\bf AI}
systems.

In the RCC theory, the \emph{Jointly Exhaustive and Pairwise
Disjoint} (JEPD) set of topological relations known as RCC8 are
identified as being of particular importance. RCC8 contains
relations ``$x$ is disconnected from $y$", ``$x$ is externally
connected to $y$", ``$x$ partially overlaps $y$", ``$x$ is equal
to $y$", ``$x$ is a tangential proper part of $y$", ``$x$ is a
non-tangential proper part of $y$", and the inverses of the latter
two relations. Interestingly, this classification of topological
relations has been independently given by Egenhofer \cite{E91} in
the context of Geographical Information Systems (GIS). Since RCC8
is JEPD, it supports a composition table. The RCC8 composition
table appears first in \cite{CCR93} and coincides with that of
\cite{E91}.

Originating in Allen's analysis of temporal relations
\cite{Allen83,Allen84}, the notion of a \emph{composition table}
({\bf CT}) has become a key technique in providing an efficient
inference mechanism for a wide class of theories
\cite{Vilain,EF91,Freksa92,RCoCu92,Rohrig,Schlieder}. It is worthy
of note that the precise meaning of a composition table depends to
some extent on the situation where it is employed.

Generally speaking, a {\bf CT} is just a mapping $CT:{\bf
Rels}\times{\bf Rels}\ra 2^{\bf Rels}$, where ${\bf Rels}$ is a
set of relation symbols \cite{DUN}. For three relation symbols
{\bf R, S} and {\bf T}, we say $\langle{\bf R,T,S}\rangle$ is a
\emph{composition triad} in {\bf CT} if ${\bf T}$ is in $CT({\bf
R,S})$. A model  of $CT$ is then a pair $\langle U,v\rangle$,
where $U$ is a set and $v$ is a mapping from ${\bf Rels}$ to the
set of binary relations on $U$ such that $\{v({\bf R}):{\bf
R}\in{\bf Rels}\}$ is a partition of $U\times U$ and $v({\bf
R})\circ v({\bf S} )\subseteq\bigcup_{{\bf T}\in{CT}({\bf
R,S})}v({\bf T})$ for all ${\bf R,S}\in{\bf Rels}$, where $\circ$
is the usual relation composition. A model $\langle U,v\rangle$ is
called \emph{consistent} if ${\bf T} \in{CT} ({\bf
R,S})\Lra(v({\bf R})\circ v({\bf S}))\cap v({\bf
T})\not=\emptyset$ for all ${\bf R,S,T}\in {\bf Rels}$
\cite{LY03a}. This means that, for any three relation symbols
${\bf T,\ R}$ and {\bf S}, ${\bf T}$ is an entry of the cell
specified by ${\bf R}$ and ${\bf S}$ if and only if there exist
three regions $a,b,c$ in $U$ such that ${\bf R}(a,b)$, ${\bf
S}(b,c)$ and ${\bf T}(a,c)$. We call a consistent model
\emph{extensional} if $v({\bf R})\circ v({\bf S})=\bigcup_{{\bf
T}\in{CT}({\bf R,S})}v({\bf T})$ for all ${\bf R,S}\in{\bf Rels}$
\cite{LY03a}. In such a model, if {\bf T} is an entry in the cell
specified by {\bf R} and {\bf S}, then whenever ${\bf T}(a,c)$
holds, there must exist some $b$ in $U$ s.t. ${\bf R}(a,b)$ and
${\bf S}(b,c)$. Note if a {\bf CT}  has an extensional model
$\langle U,v\rangle$, then by a theorem given in \cite{Jonsson82},
this {\bf CT} is the composition table of a relation algebra and
$\langle U,v\rangle$ is a representation of this relation algebra
. In what follows, when the interpretation mapping $v$ is clear
from the context, we also write $U$ for this model.

Suppose that ${\mathcal R}$ is a JEPD set of relations on a
nonempty set $U$, and ${\bf R,S}\in{\mathcal R}$. D\"{u}ntsch
\cite{DUN} defines the \emph{weak composition} of ${\bf R,S}$ as
\[{\bf R}\circ_w {\bf S}=\bigcup\{{\bf T}\in{\mathcal R}:
{\bf T}\cap{\bf R}\circ{\bf S}\neq\emptyset\}.\]In case ${\mathcal
R}$ is finite, we summarize the weak compositions in a table and
call this a \emph{weak} composition table. Note by definition, a
model $\langle U,v\rangle$ of a {\bf CT} $CT:{\bf Rels}\times{\bf
Rels}\ra 2^{\bf Rels}$ is consistent if and only if $CT$ is
precisely the weak composition table of ${\bf Rels}$ on
$U$.\footnote{What should be addressed is, although D\"{u}ntsch
call the RCC11 table \cite[Table 17]{DUN} \emph{weak}, it is not
clear or at least haven't be proven whether or not this table is
precisely the weak composition table for each RCC model.}

\vskip 3mm

Since the RCC theory entails the RCC8 {\bf CT}, each RCC model is
already a model of the RCC8 {\bf CT}. But examination of the RCC8
{\bf CT} reveals that an extensional interpretation is not
compatible with the 1st-order RCC theory. This fact is pointed out
by Bennett in \cite{Bennett98} and \cite{BIC97}. To avoid this
problem and hence construct an extensional composition table,
Bennett suggests \cite{Bennett98} to remove the universal region
from the domain of possible referents of the region constants. In
\cite{LY03a}, however, after an exhaustive investigation about
extensional interpretation of the RCC8 {\bf CT}, Li and Ying has
shown that no RCC model can be interpreted extensionally anyway.

Another way to construct an extensional composition table has also
been suggested by Bennett et al. \cite{BIC97}:
\begin{quote}
  ``One might further conjecture that by refining relations in a set
  ${\bf Rels}$ one can always arrive at a set ${\bf Rels}^\prime$
  which is more expressive than ${\bf Rels}$ and whose ${\bf CT}$
  can be interpreted extensionally."
 \end{quote}

This approach to extensional composition table relates closely to
the formalism of relation algebras initiated by Tarski
\cite{Tarski41}. Moreover, noticing that the expressiveness of
reasoning with basic operations on binary relations is equal to
the expressive power of the three variable fragment of first order
logic with at most binary relations \cite{Tarski87}, it seems
worthwhile to use methods of relation algebras to study connection
(or contact) relations in their own right. Note that Bennett's
question described above can be reformulated as the question of
determining the relation algebra generated by the connectedness
relation.

In a series of papers \cite{DUN, DWM99,DSW01,DWM01}, D\"{u}ntsch
and his colleagues study the relation-algebraic aspects of the RCC
theory systematically. They show that the contact relation algebra
contains more relations than the RCC8 relations might suggest: the
RCC8 relations has been refined to RCC10, RCC11 and RCC25, and
corresponding weak {\bf CT}s are also given. In particular, they
show \cite{DSW01} that each relation algebra generated by the
contact relation of an RCC model contains an integral algebra
$\mathfrak{A}$ with 25 atoms as a subalgebra. In the same paper,
they ask if there is an RCC model with $\mathfrak{A}$ as its
associated binary relation algebra.

Later, Mormann \cite{Mormann01} introduces the concept of
`\emph{Hole}' relation $\bH$ in RCC and shows $\bH$ and $\bC$ are
interdefinable. More importantly, he shows several RCC25 base
relations can be split by $\bH$ or some relations derived from
$\bH$, thus gives a negative answer to D\"{u}ntsch's question. He
also suggests that the hole relation $\bH$ may be used to define
various infinite families of hole relations. Interestingly, by
formalizing the concept of `\emph{Separable Proper Part}', he
shows that \cite{Mormann03}, for a large class of RCC models, the
relation algebra generated by the contact relation contains
infinitely many elements.

Similar results are also obtained in this paper. For the RCC model
${\mathfrak B}_\omega$ constructed in \cite{LY02a}, which is a
least RCC model in the sense that each RCC model contains it as a
sub-model, we show that the contact relation algebra of
${\mathfrak B}_\omega$ is not atomic complete, therefore not
finite. For the standard RCC model associated to each
${\mathbb{R}}^{n}$, we construct two strictly decreasing sequences
of `hole' relations in the associated contact relation algebra.

All these results suggest that Bennett's conjecture is not
applicable. To obtain an extensional model of the RCC8 ${\bf CT}$,
one should restrict the domain of possible regions: an RCC model
might contain too much regions. D\"{u}ntsch \cite{DUN} has shown
that the domain of closed disks of the Euclidean plane provides an
extensional model of the RCC8 {\bf CT}, namely, the relation
algebra determined by the RCC8 {\bf CT} can be represented by the
closed disk algebra. The domain of connected regions bounded by
Jordan curves, called \emph{Egenhofer} model, also provides an
extensional interpretation \cite{LY03b}. Interestingly, this model
is in a sense a maximal extensional domain of the RCC8 {\bf CT}
\cite{LY03b}. This suggests that these disk-like regions are more
suitable for the RCC8 relations. One serious problem with these
two domains of regions is neither are closed under
complementation. But, as noted by Stell \cite{Stell01b},
complement is a fundamental concept in spatial relations. These
two domains of regions are therefore too restrictive.

In \cite{DUN}, with modelling complementation in mind, D\"{u}ntsch
refines RCC8 to RCC11: the `$x$ is externally connected to $y$'
relation splits into two situations according to whether or not
$x$ is equal to $y^\prime$, the complement of $y$; the `$x$
partially overlaps to $y$' relations splits into three situations
according to whether of not $x$ is a tangential or non-tangential
proper part of $y^\prime$. The RCC11 {\bf CT} is also given
 and it ``turns out that there is a relation algebra $A$
whose composition is represented by the RCC11 table. $A$, however,
cannot come from an RCC model as Proposition 8.6 shows, and no
representation of $A$ is known" \cite{DUN}.

In the present paper, we first show that each RCC model is
consistent w.r.t. the RCC11 {\bf CT} and then an exhaustive
investigation about extensional interpretation of the RCC11 {\bf
CT} is given. In fact, we attach a superscript $^\times$ to a cell
entry in the table if and only if extensional interpretation is
impossible for this entry.

One of the main contribution of this paper is to provide an
extensional model for the RCC11 {\bf CT}. Note models of the RCC11
{\bf CT} are closed under complementation. Our model then contains
simply two kinds of regions: the closed disks and the closures of
their complements in the Euclidean plane, where two regions are
connected if they have nonempty intersection. Note this domain is
clearly a sub-domain  of the standard RCC model associated to
$\mathbb{R}^2$. We then have two methods to introduce the RCC11
relations on this domain: the first system of relations is
obtained by restriction of the RCC11 relations in the standard RCC
model associated to $\mathbb{R}^2$, the second can be defined by
the connectedness relation on this domain. Interestingly these two
systems of relations are identical. The binary relation algebra
generated by the connectedness relation, the \emph{complemented
closed disk algebra}, has 11 atoms that correspond to the RCC11
relation and the composition of this algebra is just the one
specified by the RCC11 {\bf CT}. In a word, the complemented
closed disk algebra provides a representation of the relation
algebra determined by the RCC11 {\bf CT}.

Note that hand building of composition tables even for a small
number of relations is an arduous and tedious work. Although there
are more general methods to compute composition tables (see e.g.
\cite{Kahl-Schmidt}), these methods seem not appropriate for the
present purposes. Our requirements are manifold: the method should
be applicable not only for determining the composition table, but
also for checking the consistency and extensionality of the table.
To this aim, we propose a specialized approach to reduce the
calculations: by using this approach, the work needed can be
reduced to nearly 1/8 of that needed by the cell-by-cell
verification. For example, the work need for the RCC11 {\bf CT}
has been decreased to 15 calculations of compositions, contrasting
with the $11\times 11$ cell-by-cell verifications. This approach
is also valid to other composition tables whose domain is closed
under complementation, e.g. the RCC7 weak ${\bf CT}$ and the RCC25
weak ${\bf CT}$.

The rest of the paper is arranged as follows. In next section, we
briefly summarize some basic concepts of contact relation algebras
and the RCC theory. Section 3 concerns the contact relation
algebras for certain RCC models. We first show the CRA of
${\mathfrak B}_\omega$ is not atomic complete and then construct
two infinite chains in the CRA of $n$-dimensional Euclidean space.
This fact shows that it is impossible to obtain an extensional
{\bf CT} for the RCC theory by simply refining the RCC8 relations.
The notions of dual relation set and dual generating set for RCC
relations are introduced in Section 4. Based on these notions, a
very effective approach to determine the RCC weak {\bf CT} is
introduced.  Using this approach, in Section 5, we first show each
RCC model is a consistent model of the RCC11 {\bf CT} and then
give a complete analysis of the extensionality of the RCC11 {\bf
CT}. Section 6 introduces the complemented closed disk algebra
$\mathcal{L}$ which is a representation of the relation algebra
determined by the RCC11 composition table. Summary and outlook are
given in the last section.
\section{Contact relation algebras}
In this section we summarize some basic concepts of contact
relation algebras and the RCC models. For contact relation
algebras our references are \cite{DUN,DWM99,DSW01,DUN03}, and for
RCC models
\cite{RC89,RCuCo92,CBGG97,Bennett98,Stell00b,DWM01,LY03a}.

Recall in a relation algebra (RA)
$(A,+,\cdot,-,0,1,\circ,^{^{\sim}},1^{\prime})$,
$(A,+,\cdot,-,0,1)$ is a Boolean algebra, and
$(A,\circ,^{^{\sim}},1^{\prime})$ is a semigroup with identity
$1^{\prime}$, and $a^{^{\sim\sim}}=a,\ (a\circ
b)^{^{\sim}}=b^{^{\sim}}\circ a^{^{\sim}}$. In the sequel, we will
usually identify algebras with their base set.

An important example of relation algebra is the {\it full algebra
of binary relations} on the underlying set $U$, written $(Rel(U),
\cup, \cap, -, \emptyset, U\times U, \circ, ^{^{\sim}},
1^{\prime})$, where $Rel(U)$ is the set of all binary relations on
$U$, $\circ$ is the relational composition, $^{^{\sim}}$ the
relation converse, and $1^{\prime}$ is the identity relation on
$U$. For ${\bf R}\in Rel(U)$, and $x,y,z\in U$ we usually write
$x{\bf R}y$ or ${\bf R}(x,y)$ if $(x,y)\in {\bf R}$.

Recall a subset $A$ of $Rel(U)$ which is closed under the
distinguished operations of $Rel(U)$ and contains the
distinguished constants is called an {\it algebra of binary
relations} (BRA) on $U$. A relation algebra $A$ is called {\it
representable} if it is isomorphic to a subalgebra of a product of
full algebras of binary relations,  $A$ is called {\it integral},
if $1^{\prime}$ is an atom of $A$.

\vskip 3mm

 To avoid trivialities, we always assume that
the structures under consideration have at least two elements.
Suppose that $U$ is a nonempty set of regions, and that ${\bf C}$
is a binary relation on $U$ which satisfies

(C1)\ \ {\bf C} is reflexive and symmetric,

(C2)\ \ $(\forall x,y\in U)[x=y\leftrightarrow\forall z\in U({\bf
C}(x,z)\leftrightarrow{\bf C}(y,z))]$.

\noindent D\"{u}ntsch et al. \cite{DWM99} call a binary relation
${\bf C}$ which satisfies (C1) and (C2) a \emph{contact relation};
and an RA generated by a contact relation will be called a
\emph{contact} RA (CRA). A contact relation $\bC$ on an ordered
structure $\langle U,\leq\rangle$ is said to be \emph{compatible
with} $\leq$ if $-({\bf C}\circ-{\bf C})=\;\leq$. \emph{In this
paper we only consider compatible contact relations on
orthocomplemented lattices.} Recall an \emph{orthocomplemented
lattice} \cite{Stern} is a bounded lattice $\langle
L,0,1,\vee,\wedge\rangle$ equipped with a unary complemented
operation $^\prime:L\ra L'$ such that
\[x^{\prime\prime}=x,\ \ x\wedge x^\prime=0,\ \  x\leq
y\Leftrightarrow x^\prime\geq y^\prime.\]

Suppose $L$ is an orthocomplemented lattice containing more than
four elements and \bC\ is a contact relation other than the
identity. Set $U=L\setminus\{0,1\}$. Since $1_U$ is RA definable
\cite{DUN}, we can restrict the contact relations \bC\ and other
relations definable by \bC\  on $U$.  The following relations can
then be defined from ${\bf C}$ on $U$:
\[
\begin{array}{lcl|rcl}
  {\bf DC}&=&-{\bf C}& {\bf P}&=&-({\bf C}\circ-{\bf C})\\
  1^{\prime}&=&{\bf P}\cdot{\bf P}^{\sim}&{\bf PP}&=&{\bf P}-1^{\prime}\\
  {\bf O}&=&{\bf P}^{\sim}\circ{\bf P} &{\bf PO}&=&{\bf O}\cdot-({\bf P}+{\bf P}^{\sim}) \\
  {\bf EC}&=&{\bf C}\cdot-{\bf O} &\TPP&=&{\bf PP}\cdot({\bf EC}\circ{\bf EC})\\
  {\bf NTPP}&=&{\bf PP}\cdot-\TPP&
  \sharp&=&-({\bf P}+{\bf P}^\sim)\\
  {\bf T}&=&-({\bf P}\circ{\bf P}^\sim)&
  {\bf PON}&=&{\bf O}\cdot\sharp\cdot-{\bf T}\\
  {\bf POD}&=&{\bf O}\cdot\sharp\cdot{\bf T}&{\bf ECD}&=&-{\bf
O}\cdot{\bf T}\\
{\bf ECN}&=&{\bf
  EC}\cdot-{\ECD}&\PODZ&=&\ECD\circ\NTPP\\
{\bf DN}&=&{\bf DR}-{\bf ECD}\hskip 5mm&\hskip 5mm
  \PODY&=&\POD-\PODZ.
\end{array}
\]
We have the following systems of JEPD relations on $U$ \cite{DUN}:

RCC5 relations \[{\mathcal R}_5=\{{\bf 1^{\prime}, PP,
PP^{\sim},PO, DR}\};\]

RCC7 relations \[{\mathcal R}_7=\{{\bf
1^{\prime},PP,PP^{\sim},PON,POD,ECD,DN}\};\]

RCC8 relations \[{\mathcal R}_8=\{{\bf DC, EC, PO, 1^{\prime},
TPP, NTPP, TPP^{\sim}, NTPP^{\sim}}\};\]

RCC11 relations \[{\mathcal R}_{11}=\{{\bf 1^{\prime}, TPP,
TPP^{\sim}, NTPP, NTPP^{\sim}, PON,
      PODY, PODZ, ECN, ECD, DC}\}.\]

We summarize some  characterizations of these RCC relations.

\begin{lemma}\label{lemma:char}
Suppose $L$ is an orthocomplemented lattice $L$ with $|L|>4$ and
$\bC$ is a compatible contact relation  on $L$ other than the
identity. Then for any $x,y\in U=L\setminus\{0,1\}$, we have the
following results:

{\rm(1)}\ \ $x\PON y$ {\rm iff} $x\wedge y>0$, $x\vee y<1$,
$x\wedge y^{\prime}>0$ {\rm and} $x^{\prime}\wedge y>0$;

{\rm(2)}\ \ $x\POD y$ {\rm iff} $x\wedge y>0$, $x\vee y=1$;

{\rm(3)}\ \ $x\PP y$ {\rm iff} $x<y$;

{\rm(4)}\ \ $x\ECD y$ {\rm iff} $x=y^{\prime}$;

{\rm(5)}\ \ $x\ECN y$ {\rm iff} $x<y^\prime$ {\rm and} $x\bC y$;

{\rm(6)}\ \ $x \TPP y$ {\rm iff} $x<y$ {\rm and} $x\bC
y^{\prime}$;

{\rm(7)}\ \ $x\NTPP y$ {\rm iff} $x<y$ {\rm and} $x\DC y^\prime$;

{\rm(8)}\ \ $x\PODY y$ {\rm iff} $y^\prime<x$ {\rm and}
$x^\prime\bC y^{\prime}$;

{\rm(9)}\ \ $x\PODZ y$ {\rm iff} $y^{\prime}<x$ {\rm and}
$x^\prime\DC y^{\prime}$.

\end{lemma}

In what follows, we shall often write respectively $-x$, $x+y$,
$x-y$ for $x^\prime$, $x\vee y$ and $x\wedge y^\prime$.

\subsection{Models of the RCC axioms}

The Region Connection Calculus (RCC) was originally formulated by
Randel, Cui and Cohn \cite{RCuCo92}. There are several equivalent
formulations of RCC \cite{Stell00b,DUN}, we adopt in this paper
the one in terms of Boolean connection algebra (BCA)
\cite{Stell00b}.

\begin{defi}\label{defi:bca}
{\rm A model of the RCC is a structure $\langle A,\bC\rangle$
 such that

A1. $A=\langle A;0,1,^\prime,\vee,\wedge\rangle$ is a Boolean
algebra with more than two elements.

A2. \bC\ is a symmetric and reflexive binary relation on
$A\setminus\{0\}$.

A3. $\bC(x,x^\prime)$ for any $x\in A\setminus\{0,1\}$.

A4. ${\bf C}(x,y\vee z)$ iff ${\bf C}(x,y)$ or ${\bf C}(x,z)$ for
any $x,y,z\in A\setminus\{0\}$.

A5. For  any $x\in A\setminus\{0,1\}$, there exists some $w\in
A\setminus\{0,1\}$ such that ${\bf C}(x,w)$ doesn't hold.}
\end{defi}
Stell \cite{Stell00b} calls such a construction a \emph{Boolean
connection algebra} (BCA), this conception is stronger than the
\emph{Boolean contact algebra} given by D\"{u}ntsch \cite{DUN}. In
particular, the connection in a BCA satisfies Condition (C2) and
hence is a contact relation in D\"{u}ntsch's sense.

Given a regular connected space $X$, write $\RC(X)$ for the
regular closed algebra of $X$. Then with the  standard
Whiteheadean contact (i.e. $a\bC b$ iff $a\cap b\neq\emptyset$),
$\langle \RC(X),\bC\rangle$ is a model of the RCC \cite{Gotts96}.
These models are called \emph{standard} RCC models \cite{DUN}.
Later we shall refer the standard model associated to a regular
connected space $X$ simply $\RC(X)$.

If an RCC model $A$ satisfies the following interpolation property
\cite{NW70,Biacino96} (INT for short):
\[x\NTPP y\ra\exists z(x\NTPP z\wedge z\NTPP y)\]
we call it a \emph{strong} RCC model. Standard RCC models
associated to ${\mathbb R}^n$ are strong models. There are also
RCC models which are not strong, e.g., the least RCC model
$\mathfrak{B}_\omega$ constructed in \cite{LY02a} (see Section
3.1. of this paper).

Note in general  some of the RCC11 relations, e.g. \TPP, generated
by some contact relation on an orthocomplemented lattice will be
empty. But for RCC models, all these relations are nonempty
\cite{DUN}. D\"{u}ntsch et al. \cite{DSW01} also refined the RCC11
relations and obtained 25 JEPD topological relations, namely, the
RCC25 relations. These set of relations are contained in the CRA
of each RCC model \cite{DSW01}.

\section{RCC models and their contact relation algebras}
In this section we shall show that the CRA of ${\mathfrak
B}_\omega$ is atomic incomplete and the CRA of
$\RC({\mathbb{R}}^{n})$ is infinite and hence not generated by a
finite number of atoms. This gives a negative answer to Bennett's
conjecture depicted in the introduction of this paper. A
sufficient condition for these relation algebras to be integral is
also given.
\subsection{The CRA of a least RCC model $\mathfrak{B}_\omega$}
In \cite{LY02a}, Li and Ying constructed a countable RCC model
\bomega\ which is least in the sense that each RCC model contains
a sub-model isomorphic to \bomega. We recall some basic facts
about this model.

Let $\Sigma=\{0,1\}$ and let $\Sigma^\ast$ be the set of finite
strings over $\Sigma$ with $\epsilon$ the empty string. Now for
each string $s\in \Sigma^\ast$, we associate a
left-closed-and-right-open sub-interval of $[0,1)$ as follows:
Take\\
$x_\epsilon=[0,1)$; $x_0=[0,1/2)$, $x_1=[1/2, 1)$;\\
$x_{00}=[0,1/4)$, $x_{01}=[1/4,1/2)$, $x_{10}=[1/2,3/4)$,
$x_{11}=[3/4,1)$.\\
 In general, suppose $x_s$ has been defined for a string
$s\in\{0,1\}^\ast$, we define $x_{s0}$ to be the first half
left-closed-and-right-open sub-interval of $x_s$, and $x_{s1}$ the
second half.

Write $\mathbb{B}$ the subalgebra of the powerset algebra of
$[0,1)$ generated by all $x_s$. Clearly, $\mathbb{B}$ is a
countable atomless Boolean algebra. Define a connection
$\bC_\omega$ on
$\mathbb{U}=\mathbb{B}\setminus\{\emptyset,x_\epsilon\}$ as
follows: for two regions $a,b\in\mathbb{U}$, $\bC_\omega(a,b)$ if
and only if either $a\cap b\not=\emptyset$ or there exist
$s,t,s_1\in\Sigma^\ast$ and some $n\geq 0$ with
$\{s,t\}=\{s_10\underbrace{1\cdots 1}_n,\ s_11\underbrace{1\cdots
1}_n\}$ and $x_s\subseteq a$, $x_t\subseteq b$.

Recall the following proposition in \cite{LY02a}.

\begin{prop}\label{prop:least2}

{\rm(i)}\ \ For any string $s$ and any $n\geq 1$,
$\NTPP_\omega(x_{s0},x_s)$ and $\NTPP_\omega(x_{s0},x_{s0}\cup
x_{s\underbrace{1\cdots 1}_n})$;

{\rm(ii)}\ \ For any nonempty $a\in \mathbb{B}$ and any string
$s\not=\epsilon$, $\NTPP_\omega(a,x_s)$ if and only if $a\subseteq
x_s-x_{s\underbrace{1\cdots 1}_n}$ for some $n\geq 1$.
\end{prop}

By above proposition, we have shown in \cite{LY03a} that
$\NTPP_\omega\circ\NTPP_\omega=\NTPP_\omega$ doesn't hold.
Moreover, if we write inductively
$\NTPP_\omega^{n+1}=\NTPP_\omega\circ\NTPP_\omega^n$, then, the
following theorem shows $${\bf NTPP}_\omega,\ {\bf
NTPP}_\omega^2,\ \cdots,\ {\bf NTPP}_\omega^n,\ \cdots$$ is a
strict decreasing chain.
\begin{thm}
In the RCC model \bomega, we have $\NTPP_\omega^n\not=
\NTPP_\omega^{n+1}$ for any positive integer $n$, and
$\bigcap_{n\in{\bf N}}\NTPP_\omega^n=\emptyset$.
 \end{thm}
\begin{proof}Note for any two regions $a,b\in\mathbb{U}$, if we
set $a^\ast=\bigcup\{x_{0s}:x_s\subseteq a\}$ and
$b^\ast=\bigcup\{x_{0s}:x_s\subseteq b\}$, then we have
$\NTPP_\omega(a,b)$ if and only if $\NTPP_\omega(a^\ast,b^\ast)$.
This can be easily proved by entreating the definitions of
$\bC_\omega$ and $\NTPP_\omega$.

Suppose there exist two regions $a,b\in\mathbb{U}$ with
$(a,b)\in\bigcap_{n\in{\bf N}}\NTPP_\omega^n$. By above
observation, we also have $(a^\ast,b^\ast)\in\bigcap_{n\in{\bf
N}}\NTPP_\omega^n$. Since there is a string $s=0l_1l_2\cdots l_k$
($l_i\in\{0,1\}$ for $i=1,\cdots,k$) such that $x_s\subseteq
a^\ast\subseteq b^\ast\subseteq x_0$, we have
$(x_s,x_0)\in\bigcap_{n\in{\bf N}}\NTPP_\omega^n$. We now show how
to obtain a contradiction by proving that
$(x_s,x_0)\not\in\NTPP_\omega^{k+1}$.

In general, given a string $t=t_10\underbrace{1\cdots 1}_m$ and a
region $a\in\mathbb{U}$ with $\NTPP_\omega(x_t,a)$, we claim there
exists some $p\geq 0$ such that $x_{t^\prime}\subseteq a$, where
$t^\prime=t_11\underbrace{1\cdots 1}_{m+p}$. Recall $a$ is is a
sum of finite many base regions,  $x_{s_i}$ for instance, suppose
$n$ the largest one of the lengths of these $s_i$. Then for any
string $s$ with length bigger than or equal to $n$, we have either
$x_s\subseteq a$ or $x_s\cap a=\emptyset$. Suppose for some $p$
bigger enough we have $x_{t^\prime}\cap a=\emptyset$ with
$t^\prime$ as above. Then since $x_{t^\prime}$ is externally
connected to $x_t$, we shall have $x_{t^\prime}$ is also
externally connected to $a$. This contradicts the assumption that
$\NTPP_\omega(x_t,a)$.

For a string $t$, set $\lambda(t)$ as the total number of
occurrences of 0 in $t$. The above result then can be reformulated
as follows: for a string $t=0t_1$ and a region $a\subseteq x_0$
with $\NTPP_\omega(x_t,a)$, then there exists another string
$t^\prime$ with $\lambda(t^\prime)=\lambda(t)-1$ and
$x_{t^\prime}\subseteq a$.

Now suppose there exist $a_1,a_2,\cdots, a_k, a_{k+1}=x_0$ such
that \[x_s\NTPP_\omega a_1\NTPP_\omega a_2\cdots a_{k}\NTPP_\omega
a_{k+1}=x_0,\] where $s=0l_1l_2\cdots l_k$ as above. Suppose
$\lambda(s)=m>0$. By above observation, we shall have some $s_1$
such that $\lambda(s_{i})=\lambda(s)-1$ and $x_{s_1}\subseteq
a_1$. By assumption that $a_1\NTPP_\omega a_2$ we shall have
$x_{s_1}\NTPP_\omega a_{2}$. Continuing this procedure, since
$1<m\leq k+1$, we shall obtain a string $t=0\underbrace{1\cdots
1}_{k+p}$ ($p\geq 0$) such that $x_t\subseteq a_k$, and therefore
$\NTPP(x_t,x_0)$. This cannot be true since $x_1$ is externally
connected to both $x_t$ and $x_0$. As a result we have
$(x_s,x_0)\not\in\NTPP_\omega^{k+1}$ for any $s=0l_1l_2\cdots l_k$
($l_i\in\{0,1\}$). This suggests $\bigcap_{n\in{\bf
N}}\NTPP_\omega^n=\emptyset$.

On the other hand, note if we set $s^i=0\underbrace{0\cdots 0}_i$
for $i\geq 0$, we have $$x_{s^k}\NTPP_\omega
x_{s^{k-1}}\NTPP_\omega x_{s^{k-2}}\cdots x_{s^2}\NTPP_\omega
x_{s^1}\NTPP_\omega x_{s^0}=x_0.$$ Combining these two
observations, we shall have $(x_{s^k},x_0)$ is in $\NTPP_\omega^k$
but not in $\NTPP_\omega^{k+1}$. Therefore we have shown
$\NTPP_\omega^k\not= \NTPP_\omega^{k+1}$ for any positive integer
$k$.
\end{proof}
This result shows that the CRA of countable RCC model \bomega\ is
not atomic complete, hence infinite. As a result, the weak
composition table for RCC8 relations (and all its finite
refinements  closed under inverse) \emph{cannot} be extensional
w.r.t. the RCC theory.

In next section we shall show that the CRA of standard model of
$\RC({\mathbb{R}}^{n})$ contains two strictly decreasing sequences
of relations. The proof of this result relies on a binary hole
relation defined by Mormann \cite{Mormann01}.

\subsection{Hole relations}
To show the contact relation algebra of an RCC model may contain
more relations than that given in \cite{DSW01}, Mormann introduces
the concept of Hole relation \cite{Mormann01}. This definition
captures the intuitive concept ``hole".

\begin{defi}{\rm Let $\langle A,\bC\rangle$ be a model of RCC.
Then the relation $\bH$ on $U=A\setminus\{0,1\}$ is defined as:
$\bH=\EC\cap-(\EC\circ-\bO)$.}\end{defi}

Note any region $a\in U$ is always a hole of its complement
$a^\prime$. It is natural to exclude these situations from the
definition of ``hole". Mormann also introduces the following
restricted version of hole relation \cite{Mormann01}:
$\bH^\prime=\ECN\cap\bH$, recall where $\ECN=\{ (x,y): \EC(x,y),
x\not= y^{\prime}\}$. In case $a\bH^\prime b$, we call $a$ a
non-trivial hole of $b$.

We summarize some basic properties of these two hole relations.
\begin{lemma}{\rm\cite{Mormann01}}
Let $\langle A,\bC\rangle$ be a model of RCC and set
$U=A\setminus\{0,1\}$. Then

 {\rm (1)}\ $\bH$ and $\bH^\prime$ are nonempty relations on $U$.

 {\rm (2)}\ $\bH(x,y)$ {\rm iff} $\EC(x,y)$ and $\NTPP(x,x\vee
 y)$.

 {\rm (3)}\ $\bH(x,y)$ {\rm iff}  there is some
  $z\in U$ such that $\NTPP(x,z)$ and
 $y=z-x$.

 {\rm (4)}\ The relation \ECNB\ splits as
 $\ECNB=\bH^\prime\cup{\bH^\prime}^\sim$.
\end{lemma}
The last result of above lemma shows that the contact relation
algebra of any RCC model contains a JEPD set of relations which
refines RCC25.

\begin{prop}{\rm\cite{Mormann03}}
For standard models of RCC one has $\bH^2=\bH^4$. This
implies $\bH^i=\bH^{i+2}$ for $i\geq 2$.
\end{prop}

\begin{prop}Let $\langle A,\bC\rangle$ be a model of RCC and set
$U=A\setminus\{0,1\}$. If $U=A\setminus\{0,1\}$ contains a solid
region $a$, that is, there is no region which is a non-trivial
hole of $a$, then the contact relation algebra of $A$ is not
integral.
\end{prop}
\begin{proof}
Note there are $b,c\in U$ with $b\bH^\prime c$, hence
$c{\bH^\prime}^\sim b$ and $(c,c)\in {\bH^\prime}^{\sim}\circ
\bH^\prime$. By the assumption that $a$ is a solid region, we know
that $(a,a)\not\in \bH^{\sim}\circ \bH$.  Set
$G_{1}=({\bH^\prime}^{\sim}\circ {\bH^\prime})\cap 1^{\prime}$ and
$G_{2}=1^{\prime}-G_{1}$. Then $G_1$ and $G_2$ forms a partition
of identity relation ${\bf 1^{\prime}}$. This shows that the
contact RA of $A$ is not integral. \end{proof}

\begin{prop}\label{prop:hole}
Let $\langle A,\bC\rangle$ be a model of RCC and set
$U=A\setminus\{0,1\}$. Then ${\bH^\prime}^{n}$,
$\bigcap_{i=1}^{n}{\bH^\prime}^{2i-1}$ and
$\bigcap_{i=1}^{n}{\bH^\prime}^{2i}$ are all nonempty for $n\geq
1$.
\end{prop}
\begin{proof}Note for any region $a\in U$, by the definition of RCC model,
there exists some region $b\in U$ with $a\DC b$, hence $a\NTPP
-b$. We have a sequence of regions $a_1,a_2,\cdots, a_k,\cdots$
such $a_i\NTPP a_{i+1}$ for any $i\geq 1$. Write $b_1=a_1$, and
$b_i=a_i-a_{i-1}$ for $i\geq 2$. By $a_i\NTPP a_{i+1}$, we have
$\ECN(a_i,b_{i+1})$ for $i\geq 1$. Moreover, since
$a_i=b_i+a_{i-1}$ and $a_{i-1}\NTPP a_i$, we have
$\ECN(b_i,b_{i+1})$ for $i\geq 1$.

 Define $c_i$ inductively as
follows: $c_1=a_1$, $c_i=a_i-c_{i-1}$ for $i\geq 2$. Note for
$k\geq 1$ we have $c_{2k+1}=\Sigma_{i=0}^{k}b_{2i+1}$ and
$c_{2k}=\Sigma_{i=1}^kb_{2k}$. Then we have $c_i\bH^\prime
c_{i+1}$ for $i\geq 1$. This is because that $c_i\leq a_i\NTPP
a_{i+1}=c_i+c_{i+1}$ and, by $\ECN(b_i,b_{i+1})$, we have
$\ECN(c_i,c_{i+1})$. This suggests that ${\bH^\prime}^{n}$ is
nonempty for any $n\geq 1$, one instance is $(c_1,c_{n+1})$. At
the same time, note $c_1\ECN c_{2i}$ and $c_2=b_2\leq c_{2i}$ for
$i\geq 1$, we also have $c_1\bH^\prime c_{2i}$ for  $i\geq 1$.
This suggests
$(c_1,c_{2n})\in\bigcap_{i=1}^{n}{\bH^\prime}^{2i-1}$ and
$(c_1,c_{2n+1})\in\bigcap_{i=1}^{n}{\bH^\prime}^{2i}$ for $n\geq
1$.
\end{proof}

In what follows we shall show in the CRA of standard model
$\RC({\mathbb{R}}^n)$,
\[\bH^\prime,\ \bH^\prime\cap {\bH^\prime}^3, \
\bH^\prime\cap {\bH^\prime}^3\cap{\bH^\prime}^5,\ \cdots\] and
\[{\bH^\prime}^2,\ {\bH^\prime}^2\cap {\bH^\prime}^4, \
{\bH^\prime}^2\cap {\bH^\prime}^4\cap {\bH^\prime}^6,\ \cdots\]
are two strictly decreasing sequences of relations. To this aim,
we need the following lemma.

\begin{lemma}Suppose $X$ is a regular connected space and $a,b\not=X$ are two
nonempty regular closed sets. If $a^\circ\cap b^\circ=\emptyset$,
$a\cap b\not=\emptyset$ and $a\subset(a\cup b)^\circ$, then
$\partial a\subseteq\partial b$, $\partial(a\cup b)=\partial
b-\partial a$. Moreover, $\partial a=\partial b$ if and only if
$a=b^\prime$.
\end{lemma}
\begin{proof}To begin with, note $(a\cup b)^\circ-b$ is an open
subset contained in $a$, it is also contained in $a^\circ$. For
any $p\in\partial a$, by $a\subset(a\cup b)^\circ$, we have $p\in
b$ for otherwise $p\in (a\cup b)^\circ-b\subseteq a^\circ$.
Clearly $p$ cannot be an interior point of $b$ since any
neighborhood of $p$ containing some points in $a^\circ$. As a
result we have $p\in\partial b$, hence $\partial
a\subseteq\partial b$.

Next we show $\partial(a\cup b)=\partial b-\partial a$. For any
$p\in\partial(a\cup b)$, we have $p\in X-a$  since $a\subset(a\cup
b)^\circ$. Now for any neighborhood $U$ of $p$, since $U-a$ is
also a neighborhood of $p$, we have $(U-a)\cap (a\cup
b)^\circ\not=\emptyset$, hence $U\cap b^\circ\not=\emptyset$.
Therefore $p$ is a boundary point of $b$. On the other hand, if
$p\in\partial b-\partial a$, then we have $p\not\in a$. But by
$p\not\in b^\circ=(a\cup b)^\circ-a$, we have $p\not\in (a\cup
b)^\circ$, hence $p\in\partial(a\cup b)$.

In case $\partial a=\partial b$, we have $\partial (a\cup
b)=\partial b-\partial a=\emptyset$. This holds if and only if
$a\cup b=X$ since $X$ is connected.
\end{proof}
By above lemma, note in a standard model of RCC, $a\EC b$ if and
only if $a^\circ\cap b^\circ=\emptyset$, $a\cap b\not=\emptyset$,
$a\NTPP (a+b)$ if and only if $a\subset(a\cup b)^\circ$, we have
the following
\begin{coro}\label{coro:hole}
For a standard model $\RC(X)$ of RCC, if $\bH(a,b)$, then
$\partial a\subseteq\partial b$, $\partial(a\cup b)=\partial
b-\partial a$. Moreover, $a\bH^\prime b$ only if $\partial
a\subset\partial b$.
\end{coro}
Note if $a\bH^\prime b$, then $\partial b$ can be separated into
two nonempty closed subsets, viz. $\partial a$ and $\partial
b-\partial a$. This suggests $\partial b$ contains more connected
components than $\partial a$. Upon this observation, we next show
the contact relation algebra of $\RC(\mathbb{R}^n)$ contains
infinite relations for any $n\geq 1$.

\begin{thm}
For the CRA of the standard RCC model $\RC({\mathbb{R}}^n)$, we
have
   $\bigcap_{i=1}^{k}{\bH^\prime}^{2i-1}\not= \bigcap_{i=1}^{k+1}{\bH^\prime}^{2i-1}$
  and  $\bigcap_{i=1}^{k}{\bH^\prime}^{2i}\not=\bigcap_{i=1}^{k+1}
 \bH^{2i}$ for $k\geq 1$.
 \end{thm}
\begin{proof}For $k\geq 1$, we show there exist two regions $a,b$
such that $(a,b)$ is in $\bigcap_{i=1}^{k}{\bH^\prime}^{2i-1}$ but
not in $\bigcap_{i=1}^{k+1}{\bH^\prime}^{2i-1}$. To this end, we
construct two regions $a,b$ such that
$(a,b)\in\bigcap_{i=1}^{k}{\bH^\prime}^{2i-1}$ and $\partial b$
contains $2k-1$ more connected components than $\partial a$ does.
By Corollary \ref{coro:hole}, if
$(a,b)\in\bigcap_{i=1}^{k+1}{\bH^\prime}^{2i-1}$, then $\partial
b$ should contain at least $2k+1$ more connected components than
$\partial a$ does. We shall obtain a contradiction.

We now construct two such regions. For $n=1$, we set
$b_0=(-\infty,0]$, $b_i=[i-1,i]$ for $i\geq 1$. For $k\geq 1$,
write $a_{2k-1}=\Sigma_{i=1}^{k} b_{2i-1}$ and
$a_{2k}=\Sigma_{i=0}^k b_{2i}$. Similar to the argument given in
Proposition \ref{prop:hole}, we have $a_i{\bH^\prime} a_{i+1}$ and
$a_1{\bH^\prime} a_{2i}$ for $i\geq 1$, hence $(a_1,a_{2k})\in
\bigcap_{i=1}^{k}{\bH^\prime}^{2i-1}$. Now, since $a_1$ contains
$2$ end points and $a_{2k}=b_0+b_2+\cdots+b_{2k}$ contains only
$2k+1$ end points, $(a_1,a_{2k})\not\in
\bigcap_{i=1}^{k+1}{\bH^\prime}^{2i-1}$.

For $n\geq 2$, set $a_i$ as the $n$-ball $B(o,i)$ which has radius
$i$ and is centered at $o$ for $i\geq 1$. Define $c_1=a_1$ and
$c_i=a_i-c_{i-1}$ for $i\geq 2$. Then we have
$c_i{\bH^\prime}c_{i+1}$ and $c_1{\bH^\prime}c_{2i}$, hence
$(c_1,c_{2k})\in \bigcap_{i=1}^{k}{\bH^\prime}^{2i-1}$. But since
each $\partial c_i$ contains only $i$ connected components,
$(c_1,c_{2k})\not\in \bigcap_{i=1}^{k+1}{\bH^\prime}^{2i-1}$.
\end{proof}

\section{Dual relation sets and RCC composition tables}
In this section we shall propose a specialized approach for
reducing the computational work of establishing an RCC {\bf CT}.
This approach can also be applied in determining the consistency
and extensionality of an RCC {\bf CT}.

\subsection{Dual relation set and dual generating set}

\begin{defi}{\rm Let $\langle L,\ ^\prime\rangle$ be an
orthocomplemented lattice with $|L|>4$
  and let $U=L\backslash\{0,1\}$. For two
relation ${\bf R,\ S}$ on $U$, if $(\forall x,\,y\in U)x{\bf S}y
\leftrightarrow x{\bf R}y^{\prime}$, then ${\bf S}$ is called the
{\it right dual} of ${\bf R}$ and is denoted by ${\bf R}^{d}$. If
$(\forall x,\,y\in U)x{\bf S}y\leftrightarrow x^{\prime}{\bf R}y$,
then we call ${\bf S}$ the {\it left dual} of ${\bf R}$ and denote
it by $^{d}{\bf R}$.}
\end{defi}

The right dual and the left dual are just two unitary operations
on $Rel(U)$. For any $X\subseteq Rel(U)$, we call the relation set
$X$ a {\it dual relation set} on $U$ if $X$ is closed under the
right dual and the left dual. Clearly $Rel(U)$ itself is a dual
relation set on $U$ and intersection of dual relation sets on $U$
is also dual on $U$. We define the {\it dualization} of a relation
set $X$, denoted by $d(X)$, to be the least dual relation set
containing $X$ as a subset. For a dual relation set $\mathcal{R}$,
we can find a minimal subset $\mathcal{S}$ of $\mathcal{R}$ such
that
$\mathcal{R}=\mathcal{S}\cup\mathcal{S}^d=\,^d\mathcal{S}\cup\mathcal{S}$.
We call $\mathcal{S}$ a \emph{dual generating set of
$\mathcal{R}$.}

The following lemma summarize some basic properties of these two
dual operations and can be easily checked.

\begin{lemma}Let $\langle L,\ ^\prime\rangle$ be an
orthocomplemented lattice with $|L|>4$
  and let $U=L\backslash\{0,1\}$. Suppose
$\bR$, $\bS$ are two relations on $U$. Then the following
conditions hold:

{\rm (1)}\ \ ${\bf R}^d={\bf R}\circ\ECD$, $^d{\bf
R}=\ECD\circ{\bf R}$;

{\rm (2)}\ \ ${\bf R}^{dd}={\bf R}$, $^{dd}{\bf R}={\bf R}$,
$^d(\bR^d)=(^d\bR)^d$;

{\rm (3)}\ \ ${\bf R}^{^{\sim}d^{\sim}}= {^{d}{\bf R}}$, $(^d({\bf
R}^{\sim}))^\sim={\bf R}^{d}$;

{\rm (4)}\ \ $\bR^d\cap\bS\not=\emptyset$ {\rm iff}
$\bR\cap\bS^d\not=\emptyset$;

{\rm (5)}\ \ $^d\bR\cap\bS\not=\emptyset$ {\rm iff}
$\bR\cap{^d\bS}\not=\emptyset$;

{\rm (6)}\ \ For all $x,\,y\in U$, $(x,y)\in\, {^d(\bR^d)}$ {\rm
iff} $(x^{\prime},y^{\prime})\in\bR$.
\end{lemma}

\begin{thm}\label{thm:equation} Let $\langle L,\ ^\prime\rangle$ be an
orthocomplemented lattice with $|L|>4$
  and let $U=L\backslash\{0,1\}$. Suppose \bC\ is a compatible contact
  relation of $U$ other than the identity and
${\mathcal R}$ is a JEPD set of relations in the CRA of $U$. Then
for any $\bM,\bN\in {\mathcal R}$, we always have the following
equations, where $\circ_w$ denotes the weak composition, namely
$\bM\circ_\omega \bN=\bigcup\{\bR\in{\mathcal R}:\
\bR\cap\bM\circ\bN\not=\emptyset\}$:

{\rm (1)}\ \  $(\bM\circ \bN)^{\sim}=\bN^{\sim}\circ \bM^{\sim}$;

{\rm (2)}\ \ $(\bM\circ \bN)^{d}=\bM\circ \bN^{d}$,
 $^{d}(\bM\circ\bN)= {^{d}\bM}\circ \bN$,
 $^{d}\bM\circ\bN^{d}= {^d(\bM\circ \bN)^d}$;

{\rm (3)}\ \ $(\bM\circ_{\omega}
\bN)^{\sim}=\bN^{\sim}\circ_{\omega} \bM^{\sim}$;

{\rm (4)}\ \ Suppose ${\mathcal R}$ is a dual relation set on $U$,
then $(\bM\circ_{\omega} \bN)^{d}=\bM\circ_{\omega} \bN^{d}$,
$^{d}(\bM\circ_{\omega} \bN)= {^{d}\bM}\circ_{\omega} \bN$,
$^{d}\bM\circ_\omega \bN^{d}= {^d(\bM\circ_\omega \bN)^d}$.
\end{thm}

\begin{proof} The proofs of (1), (2) and (3) are direct. For (4),
since ${\mathcal R}$ is a dual relation set on $A$, we have
$\bM\circ\ECD=\bM\circ_\omega\ECD$ and
$\ECD\circ\bM=\ECD\circ_\omega\bM$ for each $\bM\in{\mathcal R}$.
Now applying (2) and Lemma 4.1 (4), we have
$(\bM\circ_\omega\bN)^d=\bigcup\{\bR^d:\
\bR\cap\bM\circ\bN\not=\emptyset\}=\bigcup\{\bR:\
\bR^d\cap\bM\circ\bN\not=\emptyset\}=\bigcup\{\bR:\
\bR\cap(\bM\circ\bN)^d\not=\emptyset\}=\bigcup\{\bR:\
\bR\cap\bM\circ\bN^d\not=\emptyset\}=\bM\circ_\omega\bN^d$.
Similarly we have $^{d}(\bM\circ_{\omega} \bN)=
{^{d}\bM}\circ_{\omega} \bN$. The last equation now follows from
these two equations.
\end{proof}

\begin{prop}\label{prop:e=w}
Let $\langle L,\ ^\prime\rangle$ be an orthocomplemented lattice
with $|L|>4$
  and let $U=L\backslash\{0,1\}$. Suppose \bC\ is a compatible contact
  relation on $U$ other than the identity. Then for any four RCC11
  relations ${\bf R,S,T,Q}$, we have ${\bf R}\circ_\omega{\bf
  S}={\bf T}\circ_\omega{\bf Q}$ provided that ${\bf R}\circ{\bf
  S}={\bf T}\circ{\bf Q}$ holds, where $\circ_\omega$ is the weak RCC11 composition.
\end{prop}
\begin{proof}By the definition of $\circ_\omega$ and that ${\bf R}\circ{\bf
  S}={\bf T}\circ{\bf Q}$, we have
\[\begin{array}{cclcc}{\bf R}\circ_\omega{\bf
  S}&=&\{{\bf U}\in{\rm RCC11}:{\bf U}\cap{\bf R}\circ{\bf
  S}\not=\emptyset\}&&\\
  &=&\{{\bf U}\in{\rm RCC11}:{\bf U}\cap{\bf T}\circ{\bf
  Q}\not=\emptyset\}&=&{\bf T}\circ_\omega{\bf Q}.
  \end{array}\]
\end{proof}

\subsection{An approach for reducing the calculations of
weak composition table}

The above theorem suggests that, for a dual relation set
${\mathcal R}$, the work of constructing the weak composition
table can be simplified drastically.

Suppose  ${\mathcal R}$ is a dual relation set which is closed
under inverse and contains $1^{\prime}$. Let ${\mathcal S}$ be a
dual generating set of ${\mathcal R}$ which is also closed under
inverse. Denote ${\mathcal M}=\{\bR\in{\mathcal S}:\ \bR=\bR^\sim\
\mbox{and}\ \bR\not=1^{\prime}\}$ and ${\mathcal
N}=\{\bR\in{\mathcal S}:\ \bR\not=\bR^\sim\}$. Write $r,\ s,\ m,\
n$ to be the number of relations in ${\mathcal R},\ {\mathcal S},\
{\mathcal M},\ {\mathcal N}$ respectively. Then $s=m+n+1$ and
$n=2k$ for some $k\in\mathbb{N}$.

To construct the weak CT, one should compute $\bM\circ_\omega\bN$
for each $\bM,\,\bN\in{\mathcal R}$. Theorem 4.1 shows that the
work can be simplified.

There are four cases, namely, (1) $\bM,\ \bN\in{\mathcal S}$; (2)
$\bM\in{\mathcal S}$ and $\bN\not\in{\mathcal S}$; (3)
$\bM\not\in{\mathcal S}$ and $\bN\in{\mathcal S}$; (4) $\bM,\
\bN\not\in{\mathcal S}$.

 For Case
(2), since ${\mathcal S}$ is a dual generating set of
$\mathcal{R}$, we can choose $\bR\in{\mathcal S}$ such that
$\bR^d=\bN$. Then
$\bM\circ_\omega\bN=\bM\circ_\omega\bR^d=(\bM\circ_\omega\bR)^d$
by (4) of Theorem 4.1. We reduce (2) to (1). Similarly, Case (3)
and Case (4) can be reduced to (1). Therefore we only need to
check Case (1). This can be further simplified. Suppose ${\mathcal
M}=\{{\bM}_1,\ {\bM}_2,\cdots, {\bM}_m\}$ and ${\mathcal
N}=\{{\bN}_1,\ {\bN}{_1}^\sim,\ {\bN}_2,\ {\bN}{_2}^\sim,\cdots,
{\bN}_k,\ {\bN}{_k}^\sim\}$.
\begin{itemize}
  \item For $\bM,\ \bN\in{\mathcal M}$, note
  $\bM_i\circ_\omega\bM_j=(\bM_j\circ_\omega\bM_i)^\sim$.
  The work needed in this case is
  $(m\times(m+1))/2$;
  \item For $\bM\in{\mathcal M},\ \bN\in{\mathcal N}$ or
  $\bM\in{\mathcal N},\ \bN\in{\mathcal M}$, note
  $\bM_i\circ_\omega{\bN_j}^\sim=(\bN_j\circ\bM_i)^\sim$ and
  ${\bN_j}^\sim\circ_\omega{\bM_i}=(\bM_i\circ\bN_j)^\sim$. The
  work needed in this case is $2m\times k$;
  \item For $\bM,\ \bN\in{\mathcal N}$, note the following equations
  hold:
  \[\bN_i\circ_\omega\bN_j=(\bN_j^\sim\circ_\omega\bN_i^\sim)^\sim,
  \bN_i\circ_\omega\bN_j^\sim=(\bN_j\circ_\omega\bN_i^\sim)^\sim,
  \bN_i^\sim\circ_\omega\bN_j=(\bN_j^\sim\circ_\omega\bN_i)^\sim.\]
  The work needed in this case is $2k^2+k$.
\end{itemize}
Therefore the total work needed to construct the weak {\bf CT} is
$T=(m+n)(m+n+1)/2=s(s-1)/2$.

\subsection{Dual relations of RCC systems}
In this subsection we assume $\langle L,\ ^\prime\rangle$ is an
orthocomplemented lattice with $|L|>4$
  and let $U=L\backslash\{0,1\}$. We also suppose \bC\ is a compatible contact
  relation on $U$ other than the identity.
\begin{ex}{\rm RCC5, RCC8 and RCC10 are not dual on $L$. Note that {\bf
PP}$^{d}$ is not in RCC5, $\TPP^{d}$ is not in RCC8, and
$\POD^{d}$ is not in RCC10. But by Tables \ref{table:7dual} and
\ref{table:11dual}, {\rm RCC7} and {\rm RCC11} are clearly dual
relation sets.}
\end{ex}

\begin{table}\centering
\begin{tabular}{|c|c|c|c|c|c|c|c|}
  \hline
  \tiny$\bR$&\tiny$\PP$&\tiny$\PPi$&\tiny$\PON$&\tiny$\POD$&
  \tiny$\DN$&\tiny$\ECD$&\tiny$1^\prime$\\
  \hline
  \tiny$\bR^d$&\tiny$\DN$&\tiny$\POD$&\tiny$\PON$
  &\tiny$\PPi$&\tiny$\PP$&\tiny$1^\prime$&\tiny$\ECD$\\
  \tiny$^d\bR$&\tiny$\POD$&\tiny$\DN$&\tiny$\PON$&\tiny$\PP$
  &\tiny$\PPi$& \tiny$1^\prime$ & \tiny$\ECD$\\
  \tiny$^d\bR^d$ &\tiny$\PPi$&\tiny$\PP$&\tiny$\PON$&
  \tiny$\DN$&\tiny$\POD$& \tiny$\ECD$ &\tiny$1^\prime$ \\
  \hline
\end{tabular}
\caption{\label{table:7dual}Dual operations on RCC7.}
\end{table}

\begin{table}\centering
\begin{tabular}{|c|c|c|c|c|c|c|c|c|c|c|c|}
  \hline
   \tiny\bR&\tiny$\TPP$ &\tiny $\TPPi$ &\tiny $\NTPP$ &\tiny $\NTPPi$ &\tiny $\PON$
   &\tiny $\PODY$ &\tiny $\PODZ$
   &\tiny $\ECN$ &\tiny $\ECD$ &\tiny $\DC$ &\tiny $1^\prime$\\
  \hline
  \tiny$\bR^d$ &\tiny$\ECN$&\tiny$\PODY$&\tiny$\DC$&\tiny$\PODZ$&\tiny$\PON$
  &\tiny$\TPPi$&\tiny$\NTPPi$&\tiny$\TPP$&\tiny$1^\prime$&\tiny$\NTPP$&\tiny$\ECD$\\
  \tiny$^d\bR$ & \tiny$\PODY$ & \tiny $\ECN$ &\tiny $\PODZ$ & \tiny $\DC$
  & \tiny $\PON$ &\tiny$\TPP$ & \tiny $\NTPP$ &\tiny $\TPPi$&\tiny $1^\prime$
  & \tiny $\NTPPi$&\tiny $\ECD$\\
  \tiny$^d\bR^d$&\tiny $\TPPi$&\tiny$\TPP$&\tiny $\NTPPi$&\tiny $\NTPP$
  &\tiny $\PON$&\tiny $\ECN$&\tiny $\DC$&\tiny $\PODY$&\tiny $\ECD$
  &\tiny $\PODZ$&\tiny $1^\prime$ \\
  \hline
\end{tabular}
\caption{\label{table:11dual}Dual operations on RCC11.}
\end{table}
Moreover, for RCC7 and RCC11, we have
\[{\mathcal S}_{7}=\{\bf 1^{\prime},PP,PP^{\sim},PON\}\]
\noindent is a dual generating set of ${\mathcal R}_7$; and
  \[{\mathcal S}_{11}=\{\bf 1^{\prime},TPP,TPP^{\sim}, NTPP, NTPP^{\sim},PON\}\]
\noindent is a dual generating set of ${\mathcal R}_{11}$.

By Tables \ref{table:7dual} and \ref{table:11dual},
$\mathcal{S}_7$ and $\mathcal{S}_{11}$  are closed under inverse
and $^d\bR^d=\bR^\sim$ for $\bR\in{\mathcal S}_{7}$ or
$\bR\in{\mathcal S}_{11}$. Moreover, for ${\bf M, N}\in{\mathcal
S}_{7}$ or $\mathcal{S}_{11}$, by $^d{\bf M}\circ{\bf N}^d=(^d{\bf
M}^d)\circ(^d{\bf N}^d)={\bf M}^\sim\circ{\bf N}^\sim$, we have
the following
\begin{prop}
For ${\bf M, N}\in{\mathcal S}_{7}$ or $\mathcal{S}_{11}$, we have
$^d{\bf M}\circ{\bf N}^d={\bf M}^\sim\circ{\bf N}^\sim$
\end{prop}

By this proposition and Theorem \ref{thm:equation}, we have the
following equations:

(1)\ \ $\PODY\circ \PODY=\TPPi \circ \TPP$;

(2)\ \ $\PODY\circ \PODZ=\TPPi\circ \NTPP$;

(3)\ \ $\PODY\circ \ECN=\TPPi\circ \TPPi$;

(4)\ \ $\PODY\circ \DC=\TPPi\circ \NTPPi$;

(5)\ \ $\PODZ\circ \PODY=\NTPPi\circ \TPP$;

(6)\ \ $\PODZ\circ \PODZ=\NTPPi\circ \NTPP$;

(7)\ \ $\PODZ\circ \ECN=\NTPPi\circ \TPPi$;

(8)\ \  $\PODZ\circ \DC=\NTPPi\circ\NTPPi$;

(9)\ \ $\ECN\circ \PODY=\TPP\circ \TPP$;

(10)\ \ $\ECN\circ \PODZ=\TPP\circ \NTPP$;

(11)\ \ $\ECN \circ \ECN=\TPP\circ \TPPi$;

(12)\ \ $\ECN \circ \DC=\TPP\circ\NTPPi$;

(13)\ \ $\DC \circ \PODY=\NTPP\circ\TPP$;

(14)\ \ $\DC \circ \PODZ=\NTPP\circ\NTPP$;

(15)\ \ $\DC\circ \ECN=\NTPP \circ \TPPi$;

(16)\ \ $\DC \circ \DC=\NTPP\circ\NTPPi$.

Note by Proposition \ref{prop:e=w}, the relational composition
$\circ$ in above equations can be replaced by weak composition
$\circ_w$.

We now apply the approach described in Section 4.2 to RCC7 and
RCC11. Set $t=T/n^2$ to be the ratio of the work needed in our
approach to that using the cell-by-cell checking.
\begin{description}
  \item[RCC7] $r=7,\ s=4,\ m=1,\ n=2$, $T=6$ and
  $t=6/49<1/8$;
  \item[RCC11] $r=11,\ s=6,\ m=1,\ n=4$, $T=15$ and
  $t=15/121<1/8$;
\end{description}

\begin{remark} For RCC25, we assume $\langle A,\bC\rangle$ is a model of
the RCC axioms and set $U=A\setminus\{0,1\}$. We can show that
RCC25 is closed under left and right dual operations. Moreover,
set

 ${\mathcal S}_{25}=\{{\bf 1^{\prime}, TPPA, TPPA^{\sim}, TPPB,
   TPPB^{\sim}, NTPP, NTPP^{\sim}, PONXA1},\\ {\bf PONYA1, PONYA1^\sim, PONYA2,
   PONYA2^\sim, PONZ,{\bf PONYB},{\bf PONYB}^{\sim}}\}$.

\noindent Then we can show ${\mathcal S}_{25}$ is  a dual
generating set of the RCC25 relations which is closed under
inverse. Applying the approach described in Section 4.2 to RCC25,
we need to check 105 cells from the total $25\times 25$ cells.
\end{remark}

\section{Consistency and extensionality of RCC11 CT}

In this section, we consider the consistency and extensionality of
the RCC11 {\bf CT}. In what follows, we write by
$\tau_{11}:{\mathcal R}_{11}\times{\mathcal R}_{11}\ra
2^{{\mathcal R}_{11}}$ the (abstract) RCC11 {\bf CT} given in
\cite{DUN}.

\subsection{Each RCC model is a consistent model of RCC11 CT}
Although the RCC11 {\bf CT} has been established in \cite{DUN} and
D\"{u}ntsch calls this is a \emph{weak} composition table, it is
not clear or at least haven't be proven whether or not this table
is precisely the weak composition table for each RCC model.
Namely, our question is: \emph{Is each RCC model a consistent
model of the RCC11 {\bf CT}?}

In this subsection we shall show this by constructing the weak
RCC11 composition table for each RCC model. To this aim, suppose
$A$ is an RCC model and let $U=A\setminus\{0,1\}$. Applying the
approach described in Section 4.2, we can simplify the
computation. Recall ${\mathcal S}_{11}=\{\bf
1^{\prime},TPP,TPP^{\sim}, NTPP, NTPP^{\sim},PON\}$. Let
$\mathcal{M}_{11}=\{\bf PON\}$, $\mathcal{N}_{11}=\{\bf
TPP,TPP^{\sim}, NTPP, NTPP^{\sim}\}$. We need only to calculate
the 15 weak compositions appeared in Table \ref{RCC11CT-check}.

\begin{table}\centering
\begin{tabular}{|c|c|c|c|c|c|}
  \hline
  $\circ_w$ &\TPP &$\TPPi$ &\NTPP & $\NTPPi$ & \PON \\
  \hline
  \TPP & ? & ? & ? & ? & ? \\
 $\TPPi$ & ? &  & ? &  & ? \\
 \NTPP & ? &  & ? & ? & ? \\
 $\NTPPi$ &  &  &  & ? & ? \\
  \PON & &  &  &  & ? \\
  \hline
\end{tabular}
\caption{\label{RCC11CT-check}RCC11 weak compositions should be
check.}
\end{table}

\begin{prop} Suppose $A$ is an RCC model and {\bf R, S, T} are
three RCC11 relations on $U=A\setminus\{0,1\}$. Then {\bf T} is in
$\tau_{11}({\bf R,S})$ if and only if ${\bf T}\cap{\bf R}\circ{\bf
S}\not=\emptyset$.
\end{prop}
\begin{proof}Note  $\tau_{11}$ satisfies the following conditions:
$\tau_{11}({\bf R,S}^d)=(\tau_{11}({\bf R,S}))^d$,
$\tau_{11}(^d{\bf R,S})=^d\!\!(\tau_{11}({\bf R,S}))$,
$\tau_{11}(^d{\bf R,S}^d)=^d\!\!(\tau_{11}({\bf R,S}))^d$,
$\tau_{11}({\bf S^\sim,R^\sim})=(\tau_{11}({\bf R,S}))^\sim$.
Applying the approach specified in Section 4.2, we need only to
consider cells with a symbol `$?$' in Table \ref{RCC11CT-check}.
Thus there are only 15 cases to be checked. We take
$\langle{\TPPi,\TPP}\rangle$ as an example. The rest are similar.

Note if $a\TPPi b\TPP c$, then $a\wedge c\geq b>0$, hence $a$ and
$c$ cannot be related by either $\ECD$ or $\ECN$ or $\DC$.
Moreover, if $a$ is a non-tangential proper part of $c$, then so
is $b$ since $b<a$. This contradicts $b\TPP c$, hence $a\NTPP c$
cannot hold. For the same reason, $a\NTPPi c$ is also impossible.
This shows if ${\bf T}\cap\TPPi\circ\TPP\not=\emptyset$, then {\bf
T} is in $\tau_{11}({\TPPi,\TPP})$. On the other hand, suppose
{\bf T} is in $\tau_{11}({\TPPi,\TPP})$. Take $x_i$ ($i=1,2,3$)
with $\DC(x_i,x_j)$ for $i\not=j$, take $y\NTPP x_1$. For ${\bf
T}=1^\prime$, set $a=c=x_1+x_2$, $b=x_1$; For ${\bf T}=\TPP$, set
$a=x_1+x_2+x_3$, $b=x_1$, $c=x_1+x_3$; For ${\bf T}=\TPPi$, set
$a=x_1+x_3$, $b=x_1$, $c=x_1+x_3+x_2$; For ${\bf T}=\PON$, set
$a=x_1+x_2$, $b=x_1$, $c=x_1+x_3$; For ${\bf T}=\PODY$, set
$a=1-x_2-x_3-y=(x_1-y)+(1-x_1-x_2-x_3)$, $b=x_1-y$,
$c=x_1+x_2+x_3$, then $a\TPPi b\TPP c$ and $a\TPPi 1-x_1-x_2-x_3$.
Note $c^\prime=1-x_1-x_2-x_3$, we have $a\PODY c$; For ${\bf
T}=\PODZ$, take $a=x_1$, $b=x_1-y$, $c=1-y$, then $a\TPPi b\TPP c$
and $a\NTPPi c^\prime$ hold, hence $a\PODZ c$.
\end{proof}

Recall ${\bf R}\circ_w{\bf S}$ is defined to be the union of all
${\bf T}$ with ${\bf T}\cap{\bf R}\circ{\bf S}\not=\emptyset$. By
this proposition we have the following

\begin{thm} Each RCC model is a consistent model of the RCC11
{\bf CT}  $\tau_{11}:{\mathcal R}_{11}\times{\mathcal R}_{11}\ra
2^{{\mathcal R}_{11}}$. \end{thm}

\subsection{When is a composition triad extensional?}
For an RCC model $A$, or more general, a contact structure
$\langle L,\bC\rangle$ on an orthocomplemented lattice, we say a
composition triad $\langle{\bf R,T,S}\rangle$ in $\tau_{11}$ is
\emph{extensional} if ${\bf T}\subseteq{\bf R}\circ{\bf S}$. In
\cite{DUN}, D\"{u}ntsch has shown that in general the RCC11 {\bf
CT} is not extensional. As a matter of fact, he has determined for
each cell $\langle{\bf R,S}\rangle$ whether or not ${\bf
R}\circ_w{\bf S}={\bf R}\circ{\bf S}$ is true for all RCC models.
Our intention now is to give an exhaustive investigation of the
extensionality of the RCC11 table. We want to indicate, for each
triad $\langle{\bf R,S,T}\rangle$ with {\bf T} an entry in the
cell specified by the pair $\langle {\bf R,S}\rangle$, whether or
not the following condition
\[{\bf T}(x,y)\rightarrow\exists z({\bf R}(x,z)\wedge {\bf
S}(z,y))\] \noindent hold for all RCC models.\footnote{A similar
and more detailed interpretation for RCC8 {\bf CT} has been given
in \cite{LY03a}.}

To make the calculations simple, we consider only strong RCC
models, namely those models which satisfy the INT property. This
cannot be too restrictive since stand RCC models of the Euclidean
spaces are strong.

We summarize the results in Table \ref{ExtRCC11}, where a cell
entry {\bf T} (in the cell specified by $\langle{\bf R,S}\rangle$)
is attached a superscript $^\times$ if and only if ${\bf
T}\nsubseteq{\bf R}\circ {\bf S}$. In this way, we indicate for
which triad the composition is extensional, and when it need not
to be.

The following proposition suggests the approach specified in
Section 4.2 can be used to reduce the calculations.
\begin{prop}\label{prop:eqs}
Suppose $A$ is an RCC model and {\bf R,S,T} are three
RCC11 relations on $U=A\setminus\{0,1\}$. Then the following
conditions are equivalent:

\noindent
{\rm(1)}\ \  ${\bf T}\subseteq{\bf R}\circ_w{\bf S}$;\\
{\rm(2)}\ \  ${\bf T}^d\subseteq{\bf R}\circ_w {\bf S}^d$;\\
{\rm(3)}\ \  $^d{\bf T}\subseteq\;\!^d{\bf R}\circ_w{\bf S}$;\\
{\rm(4)}\ \  $^d{\bf T}^d\subseteq\;\!^d{\bf R}\circ_w {\bf
S}^d$;\\
{\rm(5)}\ \  ${\bf T}^\sim\subseteq{\bf S}^\sim\circ_w{\bf
T}^\sim$.
\end{prop}
\begin{proof}The proofs are straightforward and leave to the reader.
\end{proof}

So we need only to calculate the 15 weak compositions appeared in
Table \ref{RCC11CT-check}. The results are given in Table
\ref{ReducedRCC11CT}.
\begin{table}
\begin{tabular}{|c|c|c|c|c|c|}            \hline\hline
{\small $\circ_{w}$}&{\bf\tiny T}&{\bf\tiny Ti}&{\bf\tiny
N}&{\bf\tiny  Ni}&{\bf\tiny PN}\\                                  \hline\hline

{\bf\tiny T}

&{\tiny T,N}

&\begin{tabular}{c}{\tiny 1$^\prime$, T, Ti, DC}\\
{\tiny PN$^\times$,ECN$^{\times}$}\end{tabular}

&            {\tiny N}                                         

&\begin{tabular}{c}{\tiny Ti$^{\times}$, Ni, PN$^\times$}\\
{\tiny ECN$^\times$, DC}\end{tabular}       

&{\tiny T, N, PN, ECN, DC}\\        
                                                          \hline
{\bf\tiny Ti}

& \begin{tabular}{c}{\tiny 1$^\prime$,  T,  Ti,  PN$^{\times}$}\\
 {\tiny PDY$^{\times}$,  PDZ}
 \end{tabular} 

&           

& \begin{tabular}{c}{\tiny T$^{\times}$, N, PN$^{\times}$}\\
{\tiny PDY$^{\times}$, PDZ}\end{tabular}    

&             

&{\tiny Ti, Ni, PN, PDY, PDZ}   

\\ \hline 

{\bf\tiny N}

&  {\tiny N}                                      

&                   

&{\tiny N}              

& \begin{tabular}{c}{\tiny 1$^{\prime}$, T, Ti, N, Ni}\\
{\tiny PN, ECN, DC}\end{tabular}    

& {\tiny T, N, PN, ECN, DC}                    

\\ \hline

{\bf\tiny Ni}

&                

&                                 

&\begin{tabular}{c}{\tiny 1$^{\prime}$, T, Ti, N, Ni}\\
{\tiny PN, PDY, PDZ}\end{tabular}    

&                  

&{\tiny Ti, Ni, PN, PDY, PDZ}                    

\\ \hline 

{\bf\tiny PN}

&        

&        

&        
&        

&\begin{tabular}{c}{\tiny 1$^{\prime}$, T,  Ti, N, Ni, PN, DC}
\\{\tiny PDY, PDZ, ECN, ECD}\end{tabular}      
 \\ \hline                                                
\end{tabular}
\caption{\label{ReducedRCC11CT}{\small Reduced `extensional' {\rm
RCC11 \textbf{CT}},  where T= TPP,  N=NTPP, Ti=TPP$^{\sim}$,
Ni=NTPP$^{\sim}$,  PN=PON, PDY=PODY, PDZ=PODZ.}}
\end{table}

The verifications are similar to that given in \cite{LY03a} for
RCC8 weak \textbf{CT}. Moreover, constructions given in
\cite[Table 4, 5]{LY03a} can also be applied for the RCC11 weak
compositions. As a matter of fact, for any cell entry ${\bf R}$ in
Table \ref{ReducedRCC11CT} which is other than $\PODY,\ \PODZ,\
\ECD$, we have: (1) if a $^\times$ is attached to ${\bf
 R}$, the construction given in Table 4 of
 \cite{LY03a} for corresponding RCC8 cell entry is still
 valid; (2) if this is not the case, entreating the counter-example constructed in
 Table 5 of \cite{LY03a} will be enough.
 In particular, for strong RCC models, we have by Table 3 of
\cite{LY03a}:

$\TPP\circ\NTPP=\NTPP\circ\TPP=\NTPP\circ\NTPP=\NTPP$;

$\TPP\circ \TPP=\TPP\cup\NTPP$;

$\NTPP\circ \NTPPi=1^{\prime}\cup\TPP\cup\TPPi\cup
\NTPP\cup\PON\cup\ECN\cup\DC$;

$\NTPPi\circ \NTPP=1^{\prime}\cup\TPP\cup\TPPi
\cup\PON\cup\PODY\cup\PODZ$.

There are still 11 cell entries to be settled. For the two
negative triads, $\langle\TPPi,\PODY^\times,\TPP\rangle$ and
$\langle\TPPi,\PODY^\times,\NTPP\rangle$, take $p,q\in U$ with
$p\NTPP q$, set $a=q$, $c^\prime=q-p$, then $a\wedge c=p$. Note by
$a\TPPi c^\prime$ we have $a\PODY c$, but there cannot exist a
region $b$ with $a\TPPi b$ and $b\leq c$ since $a\wedge c=p$ is
already a non-tangential proper part of $a$. For the rest positive
composition triads, we can choose a region $b$ with the desired
property. These constructions are summarized in Table \ref{tab:+}.

\begin{table}\centering
\begin{tabular}{l|r}
  \hline

  {\small $\langle\TPPi,\PODZ,\TPP\rangle$} & {\small Set $b=a\wedge c$} \\ \hline

  {\small $\langle\TPPi,\PODZ,\NTPP\rangle$} &
  {\small Take $m$ with $c^\prime\NTPP m\NTPP a$, set $b=a-m$} \\  \hline

  {\small $\langle\TPPi,\PODY,\PON\rangle$} &
  {\small Take $m=c^\prime$, $n\NTPP (a\wedge c)$, set
  $b=m+n$}\\ \hline

  {\small $\langle\TPPi,\PODZ,\PON\rangle$} &
  {\small Take $m\NTPP c^\prime$, $n=a\wedge c$, set  $b=m+n$}\\ \hline

  {\small $\langle\NTPPi,\PODY,\PON\rangle$} &
  {\small Take $m\NTPP c^\prime$, $n\NTPP (a\wedge c)$, set $b=m+n$}\\ \hline

  {\small $\langle\NTPPi,\PODZ,\PON\rangle$} &
  {\small Take $m\NTPP c^\prime$, $n\NTPP (a\wedge c)$, set $b=m+n$}\\ \hline

  {\small $\langle\PON,\PODY,\PON\rangle$} &
  {\small  Take $m\NTPP c^\prime$, $n\NTPP a^\prime$, set  $b=m+n$}\\ \hline

  {\small $\langle\PON,\PODZ,\PON\rangle$}&
  {\small Take $m\NTPP c^\prime$, $n\NTPP a^\prime$, set  $b=m+n$}\\ \hline

  {\small $\langle\PON,\ECD,\PON\rangle$}&
  {\small Take $m\NTPP c^\prime$, $n\NTPP a^\prime$, set  $b=m+n$}\\  \hline
\end{tabular}
\caption{\label{tab:+} Positive RCC11 weak compositions and
instances of the region $b$}
\end{table}

\begin{table}
\centering
\begin{tabular}{|p{5mm}|p{7mm}|p{9mm}|p{9mm}|p{10mm}|p{10mm}|p{10mm}|p{11mm}|
p{8mm}|p{4mm}|p{9mm}|}
                                                               \hline\hline
$\circ_{\omega}$&{\bf\tiny T}&{\bf\tiny Ti}&{\bf\tiny N}
       &{\bf\tiny Ni}&{\bf\tiny PN}&{\bf\tiny PY}&{\bf\tiny PZ}&{\bf\tiny ECN}
         &{\bf\tiny ECD}&{\bf\tiny DC}\\
                                                               \hline\hline
{\bf\tiny T} &
{\begin{tabular}{l}{\tiny T}\\                     
             {\tiny N}\\
             {\tiny }
   \end{tabular}}
&
 {\begin{tabular}{l}  {\hskip -3mm}{\tiny 1$^\prime$, T, Ti}\\
             {\hskip -3mm}{\tiny DC, PN$^\times$}\\                 
             {\hskip -3mm}{\tiny ECN$^{\times}$}
   \end{tabular}}
& {\begin{tabular}{l}
           {\tiny N}\\
          {\tiny }
     \end{tabular}}                                          
&
   {\begin{tabular}{l}
              {\hskip -3mm}{\tiny Ti$^{\times}$, Ni}\\
              {\hskip -3mm}{\tiny PN$^\times$,DC}\\
             {\hskip -3mm}{\tiny ECN$^\times$}
   \end{tabular}}
&
   {\begin{tabular}{l}  {\hskip -3mm}{\tiny T,N,PN}\\              
              {\hskip -3mm}{\tiny ECN,DC}\\
               {\tiny }
   \end{tabular}}
& {\begin{tabular}{l}
           {\hskip -3mm}{\tiny T$^{\times}$,PN$^{\times}$}\\
           {\hskip -3mm}{\tiny N,PY}\\       
           {\hskip -3mm}{\tiny ECN,ECD}
\end{tabular}}
&
 {\begin{tabular}{l}
         {\hskip -3mm}{\tiny T$^{\times}$,N,PN$^{\times}$}\\          
 {\hskip -3mm}{\tiny PY$^{\times}$}\\
         {\hskip -3mm}{\tiny PZ}
 \end{tabular}}
 &
 {\begin{tabular}{l}
       {\tiny ECN}\\
        {\tiny DC}\\                   
        {\tiny }
 \end{tabular}}
& {\begin{tabular}{l}
       {\hskip -3mm}{\tiny ECN}\\                    
        {\tiny }
 \end{tabular}}
& {\begin{tabular}{l}
     {\tiny DC}\\                      
     {\tiny }
 \end{tabular}} \\                                                    \hline
{\bf\tiny Ti} &
   {\begin{tabular}{l}
          {\hskip -3mm}{\tiny 1${'}$,T}\\                            
          {\hskip -3mm}{\tiny Ti,PN$^{\times}$}\\
          {\hskip -3mm}{\tiny PY$^{\times}$}\\{\hskip -3mm}{\tiny PZ}
   \end{tabular}}
&
 {\begin{tabular}{l}
          {\hskip -3mm}{\tiny Ti,Ni}\\                 
          {\hskip -3mm}{\tiny }
   \end{tabular}}
&
 {\begin{tabular}{l}
         {\hskip -3mm}{\tiny T$^{\times}$,N}\\                   
         {\hskip -3mm}{\tiny PN$^{\times}$}\\
         {\hskip -3mm}{\tiny PY$^{\times}$}\\
         {\hskip -3mm}{\tiny PZ}
   \end{tabular}}
&
 {\begin{tabular}{l}                              
         {\tiny Ni}\\
         {\tiny }
   \end{tabular}}
&
 {\begin{tabular}{l}
        {\hskip -3mm}{\tiny Ti,Ni}\\                    
        {\hskip -3mm}{\tiny PN,PY}\\
        {\hskip -3mm}{\tiny PZ}\\
        {\hskip -3mm}{\tiny }
   \end{tabular}}
&
 {\begin{tabular}{l}                             
      {\hskip -3mm}{\tiny PY}\\
      {\hskip -3mm}{\tiny PZ}\\
       {\hskip -3mm}{\tiny }
   \end{tabular}}
&
 {\begin{tabular}{l}                             
       {\hskip -3mm}{\tiny PZ}\\
       {\hskip -3mm}{\tiny }\\
       {\hskip -3mm}{\tiny }\\
       {\hskip -3mm}{\tiny }
   \end{tabular}}
&
 {\begin{tabular}{l}                            
      {\hskip -3mm}{\tiny Ti$^\times$,Ni}\\
      {\hskip -3mm}{\tiny PN$^\times$,PY}\\
      {\hskip -3mm}{\tiny PZ}\\
      {\hskip -3mm}{\tiny ECD}
   \end{tabular}}
&
 {\begin{tabular}{l}                            
      {\hskip -4mm} {\tiny PY} \\
      {\tiny }
   \end{tabular}}
&
 {\begin{tabular}{l}                           
        {\hskip -3mm}{\tiny Ti$^{\times}$,Ni}\\
        {\hskip -3mm}{\tiny PN$^{\times}$}\\{\hskip -3mm}{\tiny ECN$^{\times}$}\\
        {\hskip -3mm}{\tiny DC}
   \end{tabular}}\\                                                 \hline 

{\bf\tiny N} & {\begin{tabular}{l}
                {\tiny N} \\                                         
                  {\tiny }
               \end{tabular}}
&  {\begin{tabular}{l}{\hskip -3mm}{\tiny T$^\times$, N}\\                  
             {\hskip -3mm}{\tiny PN$^\times$}\\
             {\hskip -3mm}{\tiny ECN$^\times$}\\
             {\hskip -3mm}{\tiny DC}
      \end{tabular}}
&{\begin{tabular}{l}{\tiny N}\\ {\tiny }\end{tabular}}              
&   {\begin{tabular}{l}{\hskip -3mm}{\tiny 1$^{\prime}$,T,Ti}\\    
     {\hskip -3mm}{\tiny N,Ni,PN}\\
   {\hskip -3mm}{\tiny ECN,DC}\\
   {\tiny }
   \end{tabular}}
&   {\begin{tabular}{l}{\hskip -3mm}{\tiny T,N,PN}\\                    
             {\hskip -3mm}{\tiny ECN}\\
             {\hskip -3mm}{\tiny DC}\\
              {\hskip -3mm}{\tiny }
   \end{tabular}}
& {\begin{tabular}{l}{\hskip -3mm}{\tiny T$^{\times}$,N}\\     
             {\hskip -3mm}{\tiny PN$^{\times}$}\\
             {\hskip -3mm}{\tiny ECN$^{\times}$}\\
             {\tiny DC}
             \end{tabular}}
&
{\begin{tabular}{l} {\hskip -3mm}{\tiny T,N,PN}\\          
             {\hskip -3mm}{\tiny PY,PZ}\\
             {\hskip -3mm}{\tiny ECN,ECD}\\
             {\hskip -3mm}{\tiny DC }
             \end{tabular}}
&       {\begin{tabular}{l}{\tiny DC}\\                                       
       {\tiny }\end{tabular}}
&     {\begin{tabular}{l}{\hskip -2mm}{\tiny DC}\\                                        
      {\tiny }\end{tabular}}
&   {\begin{tabular}{l}{\tiny DC}\\                                       
    {\tiny }\end{tabular}}
\\                                                             \hline
{\bf\tiny Ni} &

   {\begin{tabular}{l}{\hskip -3mm}{\tiny Ti$^{\times}$}\\                  
            {\hskip -3mm}{\tiny Ni}\\
            {\hskip -3mm}{\tiny PN$^{\times}$}\\
             {\hskip -3mm}{\tiny PY$^{\times}$}\\
             {\hskip -3mm}{\tiny PZ}
   \end{tabular}}
&
         {\begin{tabular}{l} {\tiny Ni}\\                                     
          {\tiny }\end{tabular}}
&

   {\begin{tabular}{l}{\hskip -3mm}{\tiny 1$^{\prime}$,T}\\    
            {\hskip -3mm}{\tiny Ti,N}\\
             {\hskip -3mm}{\tiny Ni,PN}\\
             {\hskip -3mm}{\tiny PY}\\
             {\hskip -3mm}{\tiny PZ}
   \end{tabular}}
&          {\begin{tabular}{l} {\tiny Ni}\\                                      
             {\tiny }\end{tabular}}
&
   {\begin{tabular}{l}{\hskip -3mm}{\tiny Ti,Ni}\\                     
            {\hskip -3mm}{\tiny PN}\\
             {\hskip -3mm}{\tiny PY}\\
             {\hskip -3mm}{\tiny PZ}
   \end{tabular}}
&
{\begin{tabular}{l}{\tiny PZ}\\                     
                  {\tiny }
   \end{tabular}}
&{\begin{tabular}{l}{\tiny PZ}\\                     
                     {\tiny }
   \end{tabular}}
&
     {\begin{tabular}{l}
           {\hskip -3mm}{\tiny Ti$^{\times}$,Ni}\\                 
            {\hskip -3mm}{\tiny PN$^{\times}$}\\ {\hskip -3mm}{\tiny PY$^{\times}$}\\
             {\hskip -3mm}{\tiny PZ}
   \end{tabular}}
&
{\begin{tabular}{l}{\hskip -2mm}{\tiny PZ}\\                     
                  {\tiny }
   \end{tabular}}
&
   {\begin{tabular}{l}{\hskip -3mm}{\tiny Ti,Ni}\\                      
            {\hskip -3mm}{\tiny PN,PY}\\{\hskip -3mm}{\tiny PZ,DC}\\
            {\hskip -3mm}{\tiny ECN}\\{\hskip -3mm}{\tiny ECD}
   \end{tabular}}
   \\
 \hline
{\bf\tiny PN} &

   {\begin{tabular}{l}{\hskip -3mm}{\tiny T,N}\\         
            {\hskip -3mm}{\tiny PN}\\
             {\hskip -3mm}{\tiny PY}\\
             {\hskip -3mm}{\tiny PZ}
   \end{tabular}}
&

   {\begin{tabular}{l}{\hskip -3mm}{\tiny Ti,Ni}\\                     
                   {\hskip -3mm}{\tiny PN}\\
             {\hskip -3mm}{\tiny ECN}\\
             {\hskip -3mm}{\tiny DC}
   \end{tabular}}
&

   {\begin{tabular}{l}{\hskip -3mm}{\tiny T,N}\\                     
            {\hskip -3mm}{\tiny PN}\\
             {\hskip -3mm}{\tiny PY}\\
             {\hskip -3mm}{\tiny PZ}
   \end{tabular}}
&

   {\begin{tabular}{l}{\hskip -3mm}{\tiny Ti,Ni}\\                     
               {\hskip -3mm}{\tiny PN}\\
               {\hskip -3mm}{\tiny ECN}\\
               {\hskip -3mm}{\tiny DC}
   \end{tabular}}
   &

   {\begin{tabular}{l}{\hskip -3mm}{\tiny 1$^{\prime}$,T,N,Ti}\\
                 {\hskip -3mm}{\tiny Ni,PN,PY}\\                     
             {\hskip -3mm}{\tiny PZ,DC}\\
             {\hskip -3mm}{\tiny ECN,ECD}
   \end{tabular}}
&
{\begin{tabular}{l}{\tiny T,N}\\                     
            {\tiny PN}\\
            {\tiny PY}\\
             {\tiny PZ}
   \end{tabular}}
&
{\begin{tabular}{l}{\tiny T,N}\\                     
            {\tiny PN}\\
            {\tiny PY}\\
            {\tiny PZ}
   \end{tabular}}
&
   {\begin{tabular}{l}{\tiny Ti,Ni}\\                     
             {\tiny PN}\\ {\tiny ECN}\\
             {\tiny DC}
   \end{tabular}}
&
{\begin{tabular}{l}{\hskip -2mm}{\tiny PN}\\                     
                      \end{tabular}}
&
   {\begin{tabular}{l}{\tiny Ti,Ni}\\                     
             {\tiny PN}\\ {\tiny ECN}\\
             {\tiny DC}
   \end{tabular}}

\\                 
 \hline
{\bf\tiny PY} &

   {\begin{tabular}{l}                          
            {\hskip -3mm}{\tiny PY}\\
             {\hskip -3mm}{\tiny PZ}\\{\hskip -3mm}{\tiny }
   \end{tabular}}
&

   {\begin{tabular}{l}{\hskip -3mm}{\tiny Ti$^{\times}$,Ni}\\             
             {\hskip -3mm}{\tiny PN$^{\times}$}\\ {\hskip -3mm}{\tiny PY,ECN}\\
             {\hskip -3mm}{\tiny ECD}
   \end{tabular}}
&

   {\begin{tabular}{l}{\tiny PZ}\\                     

   \end{tabular}}
&

   {\begin{tabular}{l}{\hskip -3mm}{\tiny Ti$^{\times}$,Ni}\\      
             {\hskip -3mm}{\tiny PN$^{\times}$}\\
             {\hskip -3mm}{\tiny ECN$^{\times}$}\\
             {\hskip -3mm}{\tiny DC}
   \end{tabular}}
   &

   {\begin{tabular}{l}
              {\hskip -3mm}{\tiny Ti,Ni}\\                     
             {\hskip -3mm}{\tiny PN}\\
             {\hskip -3mm}{\tiny PY}\\
             {\hskip -3mm}{\tiny PZ}
   \end{tabular}}
&
{\begin{tabular}{l}{\hskip -3mm}{\tiny 1$^{\prime}$,T,Ti}\\     
            {\hskip -3mm}{\tiny PN$^{\times}$}\\
             {\hskip -3mm}{\tiny PY$^{\times}$}\\
               {\hskip -3mm}{\tiny PZ}
   \end{tabular}}
&
{\begin{tabular}{l}{\hskip -3mm}{\tiny T$^{\times}$,N}\\                     
            {\hskip -3mm}{\tiny PN$^{\times}$}\\  {\hskip -3mm}{\tiny PY$^{\times}$}\\
             {\hskip -3mm}{\tiny PZ}
   \end{tabular}}
&
   {\begin{tabular}{l}{\tiny Ti}\\ {\tiny Ni}\\           
                       {\tiny }
   \end{tabular}}
&
{\begin{tabular}{l}{\hskip -2mm}{\tiny Ti}\\                     
                    {\tiny }
   \end{tabular}}
&
   {\begin{tabular}{l}{\tiny Ni}\\                     
                      {\tiny }
   \end{tabular}}

\\                                                                     
 \hline
{\bf\tiny PZ} &

   {\begin{tabular}{l}{\hskip -3mm}{\tiny PZ}\\                     
                      {\hskip -3mm}{\tiny }
   \end{tabular}}
&

   {\begin{tabular}{l}{\hskip -3mm}{\tiny Ti$^{\times}$}\\             
             {\hskip -3mm}{\tiny Ni,PN$^{\times}$}\\
             {\hskip -3mm}{\tiny PY$^{\times}$}\\ {\hskip -3mm}{\tiny PZ}
   \end{tabular}}
&

   {\begin{tabular}{l}                                 
                         {\tiny PZ}\\ {\tiny }
   \end{tabular}}
&

   {\begin{tabular}{l}{\hskip -3mm}{\tiny Ti,Ni,PN}\\         
             {\hskip -3mm}{\tiny PY,PZ}\\
             {\hskip -3mm}{\tiny ECN,ECD}\\ {\hskip -3mm}{\tiny DC }
   \end{tabular}}
   &

   {\begin{tabular}{l}
              {\hskip -3mm}{\tiny Ti,Ni}\\                     
             {\hskip -3mm}{\tiny PN}\\
             {\hskip -3mm}{\tiny PY }\\
             {\hskip -3mm}{\tiny PZ}
   \end{tabular}}
&
{\begin{tabular}{l}{\hskip -3mm}{\tiny Ti$^{\times}$,Ni}\\         
            {\hskip -3mm}{\tiny PN$^{\times}$}\\
            {\hskip -3mm}{\tiny PY$^{\times}$}\\
             {\hskip -3mm}{\tiny PZ}
   \end{tabular}}
&
{\begin{tabular}{l}{\hskip -3mm}{\tiny 1$^{\prime}$,T}\\
             {\hskip -3mm}{\tiny Ti,N}\\              
              {\hskip -3mm}{\tiny Ni,PN}\\
             {\hskip -3mm}{\tiny PY,PZ }
   \end{tabular}}
&
   {\begin{tabular}{l}{\tiny Ni}\\                     
                     {\tiny }
   \end{tabular}}
&
{\begin{tabular}{l}{\hskip -2mm}{\tiny Ni}\\                     
                    {\tiny }
   \end{tabular}}
&
   {\begin{tabular}{l}{\hskip -3mm}{\tiny Ni}\\                
                      {\hskip -3mm}{\tiny }
   \end{tabular}}

\\                            
 \hline
{\bf\tiny ECN} &

   {\begin{tabular}{l}{\hskip -3mm}{\tiny T$^{\times}$,N}\\  
            {\hskip -3mm}{\tiny PN$^{\times}$}\\ {\hskip -3mm}{\tiny PY}\\
             {\hskip -3mm}{\tiny ECN}\\
             {\hskip -3mm}{\tiny ECD}
   \end{tabular}}
&

   {\begin{tabular}{l}                     
                         {\tiny ECN}\\ {\tiny DC}\\ {\tiny }
   \end{tabular}}
&

   {\begin{tabular}{l}                            
                    {\hskip -3mm}{\tiny T$^{\times}$,N}\\
                    {\hskip -3mm}{\tiny PN$^{\times}$}\\
                     {\hskip -3mm}{\tiny PY$^{\times}$}\\
                    {\hskip -3mm}{\tiny PZ}
   \end{tabular}}
&

   {\begin{tabular}{l}                     
             {\tiny DC}\\ {\tiny }
   \end{tabular}}
   &

   {\begin{tabular}{l}{\hskip -3mm}{\tiny T,N}\\        
                        {\hskip -3mm}{\tiny PN}\\
                         {\hskip -3mm}{\tiny ECN}\\
             {\hskip -3mm}{\tiny DC}

   \end{tabular}}
&
{\begin{tabular}{l}{\tiny T}\\ {\tiny N}\\                     
                   {\tiny }
   \end{tabular}}
&
{\begin{tabular}{l}{\tiny N}\\                     
                 {\tiny }
   \end{tabular}}
&
   {\begin{tabular}{l}{\hskip -3mm}{\tiny 1$^{\prime}$,T,Ti}\\      
             {\hskip -3mm}{\tiny PN$^{\times}$}\\
             {\hskip -3mm}{\tiny ECN$^{\times}$}\\
             {\hskip -3mm}{\tiny DC} {\tiny }
   \end{tabular}}
&
{\begin{tabular}{l}{\hskip -2mm}{\tiny T}\\                     
                    {\tiny }
   \end{tabular}}
&
   {\begin{tabular}{l}{\hskip -3mm}{\tiny Ti$^{\times}$ Ni}\\   
             {\hskip -3mm}{\tiny PN$^{\times}$}\\
             {\hskip -3mm}{\tiny ECN$^{\times}$}\\
             {\hskip -3mm}{\tiny DC}
   \end{tabular}}

\\                 
 \hline

{\bf\tiny ECD} &

   {\begin{tabular}{l}                          
            {\hskip -3mm}{\tiny PY }\\
   \end{tabular}}
&

   {\begin{tabular}{l}                     
                          {\hskip -3mm}{\tiny ECN }
   \end{tabular}}
&

   {\begin{tabular}{l}                            
                    {\hskip -3mm}{\tiny PZ }
   \end{tabular}}
&

   {\begin{tabular}{l}                     
             {\hskip -3mm}{\tiny DC }
   \end{tabular}}
   &

   {\begin{tabular}{l}              
             {\hskip -3mm}{\tiny PN }
   \end{tabular}}
&
{\begin{tabular}{l}{\hskip -3mm}{\tiny T }\\                     

   \end{tabular}}
&
{\begin{tabular}{l}{\hskip -3mm}{\tiny N }\\                     
   \end{tabular}}
&
   {\begin{tabular}{l}{\hskip -3mm}{\tiny Ti }\\      
   \end{tabular}}
&
{\begin{tabular}{l}{\hskip -2mm}{\tiny 1$^{\prime}$ }\\                     
   \end{tabular}}
&
   {\begin{tabular}{l} {\hskip -3mm}{\tiny Ni }   
   \end{tabular}}
\\              \hline      

{\bf\tiny DC} &

   {\begin{tabular}{l}{\hskip -3mm}{\tiny T$^{\times}$, N}\\                     
            {\hskip -3mm}{\tiny PN$^{\times}$}\\
             {\hskip -3mm}{\tiny ECN$^{\times}$}\\
             {\hskip -3mm}{\tiny DC}
   \end{tabular}}
&

   {\begin{tabular}{l}                     
                          {\tiny  DC}\\ {\tiny }
   \end{tabular}}
&

   {\begin{tabular}{l}                            
                    {\hskip -3mm}{\tiny T,N,PN}\\
                    {\hskip -3mm}{\tiny PY,PZ}\\
                    {\hskip -3mm}{\tiny ECN,ECD}\\
                    {\hskip -3mm}{\tiny DC }
   \end{tabular}}
&

   {\begin{tabular}{l}                     
            {\tiny DC}\\ {\tiny }
   \end{tabular}}
   &

   {\begin{tabular}{l}{\hskip -3mm}{\tiny T,N}\\        
             {\hskip -3mm}{\tiny PN}\\
             {\hskip -3mm}{\tiny ECN}\\
             {\hskip -3mm}{\tiny DC} {\tiny }
   \end{tabular}}
&
{\begin{tabular}{l}{\tiny N}\\                     
                   {\tiny }
   \end{tabular}}
&
{\begin{tabular}{l}{\tiny N}\\                     
                 {\tiny }
   \end{tabular}}
&
   {\begin{tabular}{l}{\hskip -3mm}{\tiny T$^{\times}$,N}\\ 
             {\hskip -3mm}{\tiny PN$^{\times}$}\\
             {\hskip -3mm}{\tiny ECN$^{\times}$}\\
             {\hskip -3mm}{\tiny DC}
   \end{tabular}}
&
{\begin{tabular}{l}{\hskip -2mm}{\tiny N}\\                     
                   {\tiny }
   \end{tabular}}
&
   {\begin{tabular}{l}{\hskip -3mm}{\tiny 1$^{\prime}$,T,Ti}\\          
            {\hskip -3mm}{\tiny N, Ni}\\
             {\hskip -3mm}{\tiny PN,ECN}\\
             {\hskip -3mm}{\tiny DC }
   \end{tabular}}

\\                 
 \hline                                                
\end{tabular}
\caption{\label{ExtRCC11}Extensionality of {\rm RCC11}
\textbf{CT}, where T=TPP, N=NTPP, Ti=TPP$^{\sim}$,
Ni=NTPP$^{\sim}$, PN=PON, PY=PODY, PZ=PODZ.}
\end{table}

\section{Complemented closed disk algebra}
This section shall provide  a representation
for the relation algebra determined by the RCC11 {\bf CT}. Recall
$\RC({\mathbb R}^2)$, the standard RCC model associated to the
Euclidean plane,
 contains all regular closed subsets of ${\mathbb R}^2$, and two (nonempty)
 regions are said to be connected provided that they have
 nonempty intersection.

 Our domain of regions, denoted by $D$,  is a sub-domain of $\RC({\mathbb
R}^2)$ and  contains two classes of
 regions: the closed disks and their
complements in $\RC({\mathbb R}^2)$. We denote by $D_1$ the class
of closed disks, by $D_2$ the class of their complements and call
for convenience regions in $D_2$ \emph{complement disks}. Define a
binary relation $\bC$ on $D$ as follows: for two regions $a,b\in
D$, $a\bC b$ if $a\cap b\not=\emptyset$. Clearly this relation is
a contact relation on $U$. In contrast with the closed disk
algebra for RCC8 table given in
 \cite{DWM99,DUN}, we call the
contact relation algebra on this domain the \emph{complemented
closed disk algebra}, written $\mathcal{L}$. In what follows we
shall show this CRA is finite and contains RCC11 as its atoms, and
it is indeed a representation of the relation algebra determined
by the RCC11 {\bf CT}. Interestingly, the RCC11 relations on this
domain can be equivalently determined by the 9-intersection
principle of Egenhofer and Herring \cite{EH91}.

\subsection{Topological characterization of RCC11 relations in $\mathcal{L}$}

Write $L=D\cup\{\emptyset,\mathbb{R}^2\}$. Then $L$ with the usual
inclusion ordering is an orthocomplemented lattice. Based on the
contact relation \bC\ on $D$, we can define RCC11 relations on $D$
(see Section 2 of this paper).

The following theorem gives a topological characterization of
these relations:

\begin{thm}\label{thm:RCC11}
The RCC11 relations on $D$ has the following characterization:

{\rm(1) $x1^{\prime}$ y iff $x=y$;

(2) $x \TPP y$ iff $x\subseteq y$, $x\not= y$ and $\partial x\cap
\partial y\not=\emptyset$;

(3) $x \TPPi y$ iff $x\supseteq y$, $x\not= y$ and $\partial x\cap
\partial y\not=\emptyset$;

(4) $x\NTPP y$ iff $x\subseteq y$, $x\not=y$ and $\partial x\cap
\partial y=\emptyset$;

(5) $x\NTPPi y$ iff $x\supseteq y$, $x\not= y$ and $\partial x\cap
\partial y=\emptyset$;

(6) $x\PON y$ iff $x^{\circ}\cap y^{\circ}\not=\emptyset$,
$x\not\subseteq y$, $y\not\subseteq x$, and $x\cup
y\not={\mathbb{R}}^2$;

(7) $x\PODY y$ iff  $x^{\circ}\cap y^{\circ}\not=\emptyset$,
$\partial x\cap \partial y\not=\emptyset$ and $x\cup
y={\mathbb{R}}^2$;

(8) $x\PODZ y$ iff  $x^{\circ}\cap y^{\circ}\not=\emptyset$,
$\partial x\cap \partial y=\emptyset$ and $x\cup
y={\mathbb{R}}^2$;

(9) $x\ECN y$ iff $x^{\circ}\cap y^{\circ}=\emptyset$, $x\cap
y\not=\emptyset$ and $x\cup y\not={\mathbb{R}}^2$;

(10) $x\ECD y$ iff $x^{\circ}\cap y^{\circ}=\emptyset$, $x\cap
y\not=\emptyset$ and $x\cup y={\mathbb{R}}^2$;

(11) $x\DC y$ iff $x\cap y=\emptyset$.}
\end{thm}
\begin{proof}The proofs are routine and leave to the
reader.\end{proof}

\begin{figure}\centering
\includegraphics{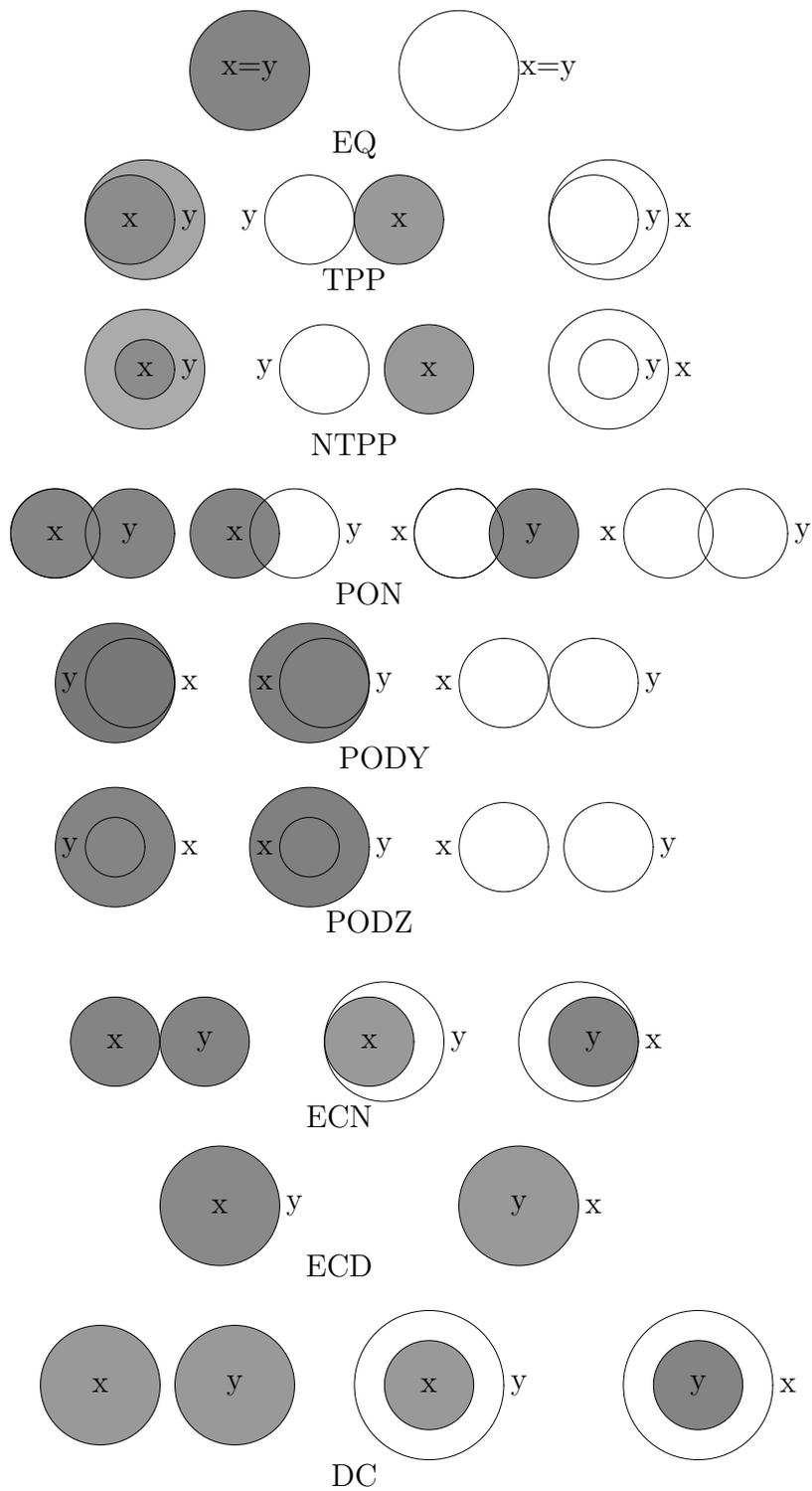}
\caption{Illustration of RCC11 relations in the complemented
closed disk algebra.} \label{fig:rcc11}
\end{figure}

From this theorem we know that these relations on $D$ are
precisely the restrictions of corresponding RCC11 relations in
$\RC({\mathbb R}^2)$ to $D$. The corresponding configurations are
illustrated in Figure~\ref{fig:rcc11}, where we figure closed
disks as shaded circles and their complements as hollowed circles.
Because $\TPP^{\sim}$ and $\NTPP^{\sim}$ are inverse relations of
$\TPP$ and $\NTPP$ respectively, we give 9 figures of the 11
relations.

\subsection{9-Intersection relations on $D$}

Interestingly, these 11 relations on $D$ can be  classified by the
9-intersection principle posed by Egenhofer and Herring
\cite{EH91}. According to the  9-intersection principle, the
binary topological relation ${\bf R}$ between two regions, $x$ and
$y$, is based upon the intersections of $x$'s interior
($x^{\circ})$, boundary ($\partial x$), and exterior ($x^{-}$)
with $y$'s interior ($y^{\circ}$), boundary ($\partial y$), and
exterior ($y^{-}$), which is concisely represented as a $3\times
3$-matrix
\[
 \left(
\begin{array}{ccc}
  x^{\circ}\cap y^{\circ} & x^{\circ}\cap \partial y & x^{\circ}\cap y^{-}\\
  \partial x\cap y^{\circ} & \partial x\cap \partial y & \partial x\cap y^{-}\\
  x^{-}\cap y^{\circ} & x^{-}\cap \partial y & x^{-}\cap
  y^{-}
\end{array}
\right).
\]
For the 9-intersection mode, the \emph{content} of the nine
intersections was identified as a simple and most general
topological invariant, it characterizes each of the nine
intersections by a value \emph{empty} ($0$) or \emph{nonempty}
($1$). The sequence of the nine intersections, from left to right
and from top to bottom, will always be in ordering $\langle
interior, boundary, exterior\rangle$.

The nine empty/nonempty intersections describe a set of relations
that provides a complete coverage--any set is either empty or not
empty and \emph{tertium non datur}. Furthermore, these relations
are \emph{Jointly Exhaustive and Pairwise Disjoint} (JEPD).

By applying the 9-intersection principle to our domain of regions
$D$, we find there are 11 JEPD relations on $D$. Moreover, these
11 relations are just the same as the RCC11 relations on $D$. This
fact follows from the topological characterization
 of the relations. We describe these RCC11 relations by $3\times 3$ matrixes as follows:
\[\begin{array}{lll}
    1{'}=\left(
              \begin{array}{rrr}
              1 & 0 & 0\\
              0 & 1 & 0\\           
              0 & 0 & 1
              \end{array}
              \right),
& \TPP=\left(
              \begin{array}{ccc}
              1 & 0 & 0\\
              1 & 1 & 0\\      
              1 & 1 & 1
              \end{array}
              \right),
& \TPPi=\left(
              \begin{array}{rrr}
              1 & 1  & 1\\
              0 & 1 & 1\\        
              0 & 0 & 1
              \end{array}
              \right),\\
\NTPP= \left(
               \begin{array}{rrr}
               1 & 0 & 0\\
               1 & 0 & 0\\           
               1 & 1 & 1
               \end{array}
               \right),
& \NTPPi= \left(
               \begin{array}{rrr}
               1 & 1 & 1\\
               0 & 0 & 1\\            
               0 & 0 & 1
               \end{array}
               \right),
& \PON=\left(
               \begin{array}{rrr}
               1 & 1 & 1\\
               1 & 1 & 1\\     
               1 & 1 & 1
               \end{array}
               \right),\\
 \PODY=\left(
               \begin{array}{rrr}
               1 & 1 & 1\\
               1 & 1 & 0\\          
               1 & 0 & 0
               \end{array}
               \right),
& \PODZ=\left(
               \begin{array}{rrr}
               1 & 1 & 1\\
               1 & 0 & 0\\             
               1 & 0 & 0
               \end{array}
               \right),
& \ECN= \left(
               \begin{array}{rrr}
               0 & 0 & 1\\
               0 & 1 & 1\\          
               1 & 1 & 1
               \end{array}
               \right),\\
\ECD=\left(
               \begin{array}{rrr}
               0 & 0 & 1\\
               0 & 1 & 0\\              
               1 & 0 & 0
               \end{array}
               \right),
& \DC=\left(
               \begin{array}{rrr}
               0 & 0 & 1\\
               0 & 0 & 1\\           
               1 & 1 & 1
              \end{array}
              \right).&
\end{array}\]

\subsection{The composition of the complemented closed disk algebra}

Now we shall show that the composition operation of $\mathcal{L}$
is precisely that one specified by the  RCC11 {\bf CT}. What we
should do is to indicate, for each triad $\langle{\bf
R,T,S}\rangle$ with {\bf T} an entry in the cell specified by the
pair $\langle {\bf R,S}\rangle$, whether or not the following
condition hold
\[{\bf T}(x,y)\rightarrow(\exists z\in D)({\bf R}(x,z)\wedge {\bf
S}(z,y)).\]

Note the approach described in Section 4.2 is also valid for the
present purpose. This is since (i) RCC11 relations on $D$ is a
dual relation set which contains $1^\prime$ and is closed under
inverse; (ii) ${\mathcal S}_{11}=\{1^{\prime}, {\bf TPP,
TPP^{\sim}, NTPP, NTPP^{\sim}, PON}\}$ is a dual generating set
which is also closed under inverse; (iii) Proposition
\ref{prop:eqs} is still valid for $\mathcal{L}$. As a result, we
need only to calculate the 15 compositions appeared in Table
\ref{RCC11CT-check}.

To begin with, we first show the $\NTPP$ relation on $D$ satisfies
the interpolation property.

\begin{lemma}Given any two regions $a,c$ in $D$ with $a\NTPP c$,
there exists another region $b\in D$ with $a\NTPP b\NTPP c$.
\end{lemma}
\begin{proof}By the topological characterization of the $\NTPP$
relation given in Theorem \ref{thm:RCC11}, we know that $a\NTPP c$
if and only if $a\subset c^\circ$. There are three cases:

Case I: $a$, $c$ are closed disks. In this case, $\partial a$  and
$\partial c$ are two non-tangential circles and $\partial a$ is
inside $\partial c$. Then we can find another circle $B$ between
these two circles. Taking $b$ as the closed disk bounded by $B$,
then $b$ satisfies the desired property.

Case II: $a$, $c$ are complement disks. In this case, $\partial a$
and $\partial c$  are two non-tangential circles and $\partial c$
is inside $\partial a$. Then we can find another circle $B$
between these two circles. Taking $b$ as the complement disk
bounded by $B$, then $b$ satisfies the desired property.

Case III: $a$ is a closed disk and $c$ is a complement disk,
$\partial a$  and $\partial c$ are two separated circles and the
distance between them is non-zero. Then we can find another circle
$B$ such that $\partial a$ is inside $B$ and $B$ is separated from
$\partial c$. Taking $b$ as the closed disk bounded by $B$, then
$b$ satisfies the desired property.
\end{proof}
\begin{prop}In the complemented closed disk algebra $\mathcal{L}$,
 the following equations $\NTPP\circ \NTPP=\NTPP$, $\TPP\circ
\NTPP=\NTPP$ and $\NTPP\circ\TPP=\NTPP$ hold.
\end{prop}
\begin{proof}Note the ``$\subseteq$" part of these equations follow directly
from the definitions and the first equation is then clear by above
lemma.

For the second equation, suppose $a\NTPP c$ in $D$, we want to
find $b$ such that $a\TPP b\NTPP c$. There are three cases:

Case I: $a$, $c$ are closed disks. In this case, $\partial a$  and
$\partial c$ are two non-tangential circles and $\partial a$ is
inside $\partial c$. Then we can find another circle $B$ such that
$\partial a$ is internally tangent to $B$ and $B$ is inside the
circle  $\partial c$. Taking $b$ as the closed disk bounded by
$B$, then $b$ satisfies the desired property.

Case II: $a$, $c$ are complement disks. In this case, $\partial a$
and $\partial c$  are two non-tangential circles and $\partial c$
is inside $\partial a$. Then we can find another circle $B$ such
that $B$ is internally tangent to $\partial a$ and $\partial c$ is
inside $B$. Taking $b$ as the complement disk bounded by $B$, then
$b$ satisfies the desired property.

Case III: $a$ is a closed disk and $c$ is a complement disk,
$\partial a$  and $\partial c$ are two separated circles and the
distance between them is non-zero. Then we can find another circle
$B$ such that $\partial a$ is internally tangent to $B$ and $B$ is
separated from $\partial c$. Taking $b$ as the closed disk bounded
by $B$, then $b$ satisfies the desired property.

The proof of the last equation is similar.
\end{proof}

The following proposition proves the remainder 12 equations in
CCA.

\begin{figure}\centering
\includegraphics[height=2.2cm]{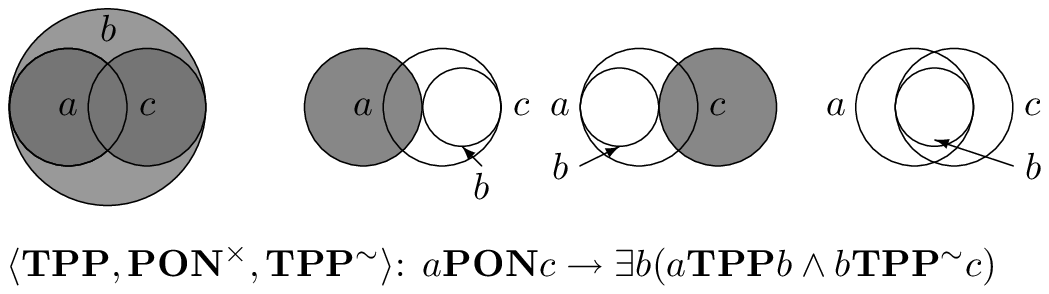}
\includegraphics[height=2.1cm]{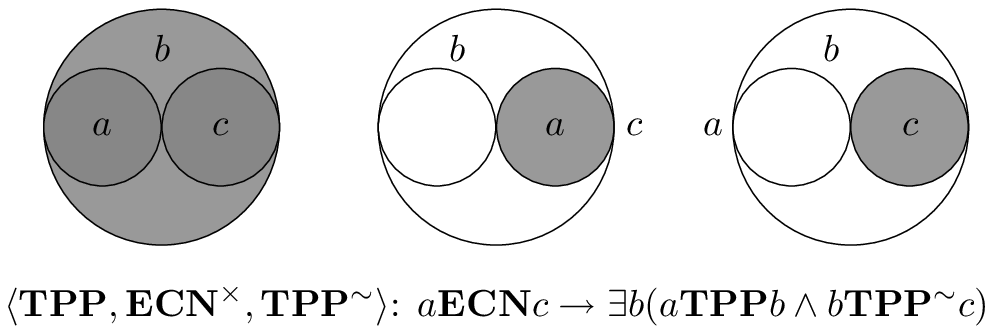}
\includegraphics[height=2.2cm]{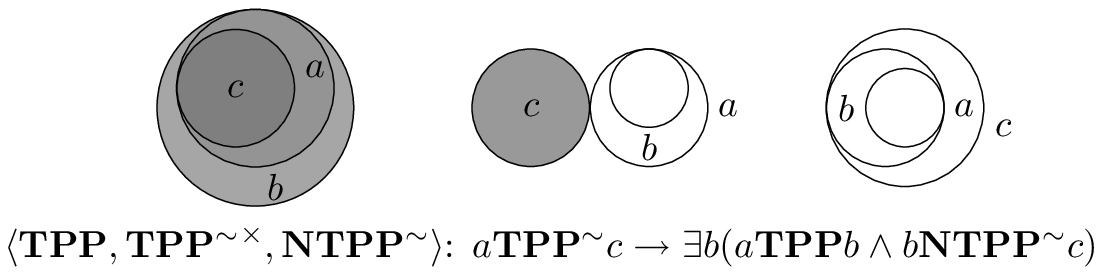}
\includegraphics[height=2.3cm]{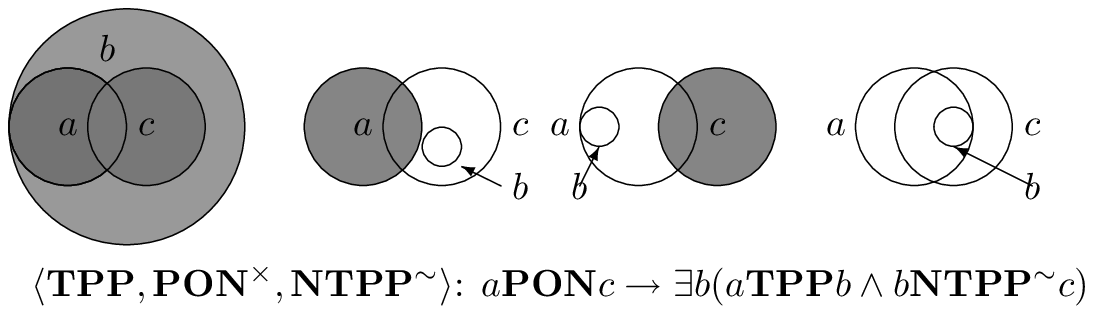}
\includegraphics[height=2.2cm]{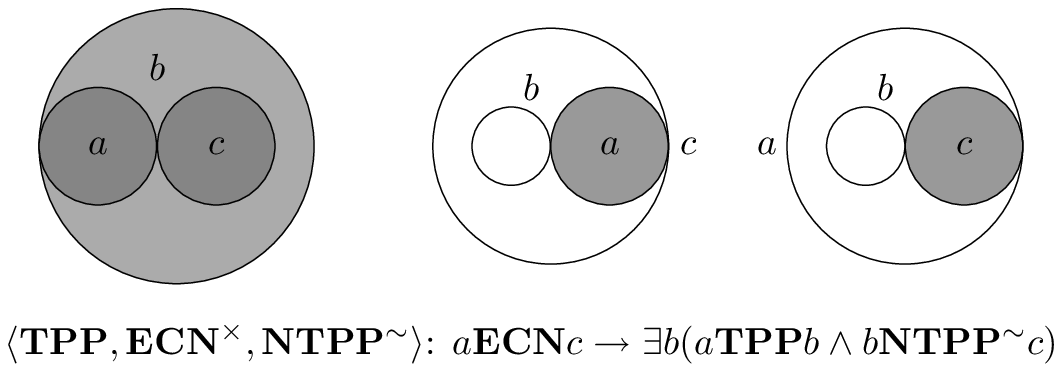}
\includegraphics[height=2.0cm]{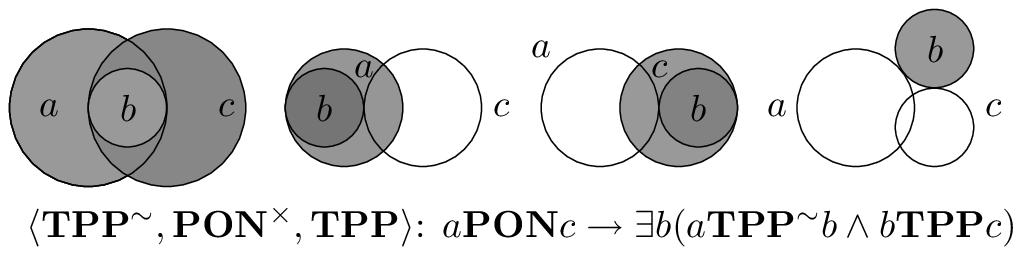}
\includegraphics[height=2.0cm]{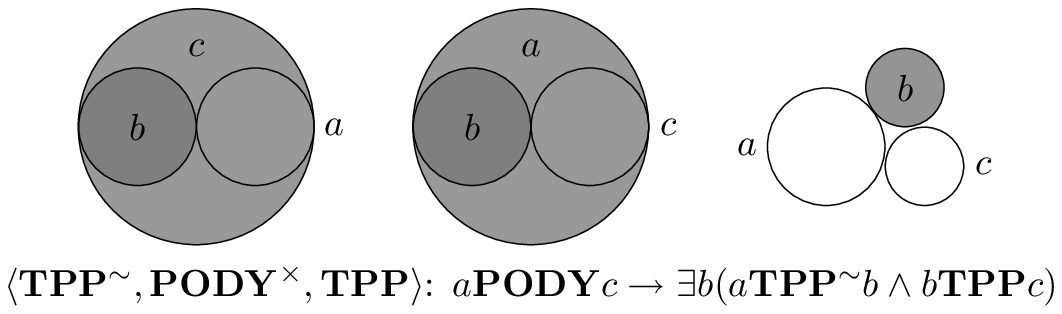}
\includegraphics[height=2.2cm]{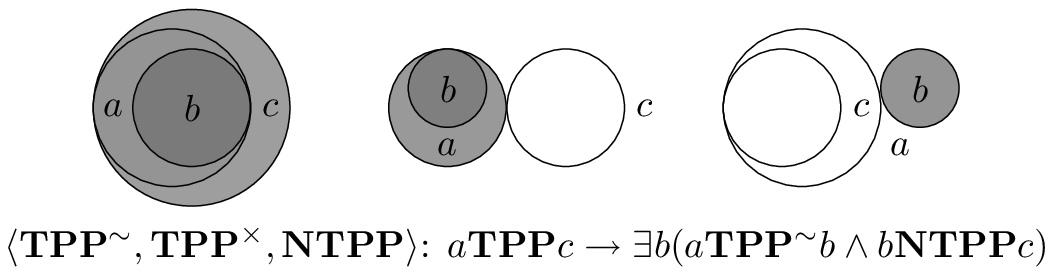}
\includegraphics[height=2.2cm]{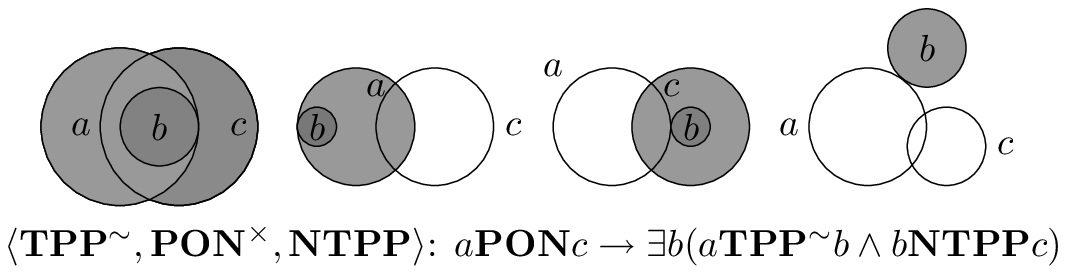}
\includegraphics[height=2.0cm]{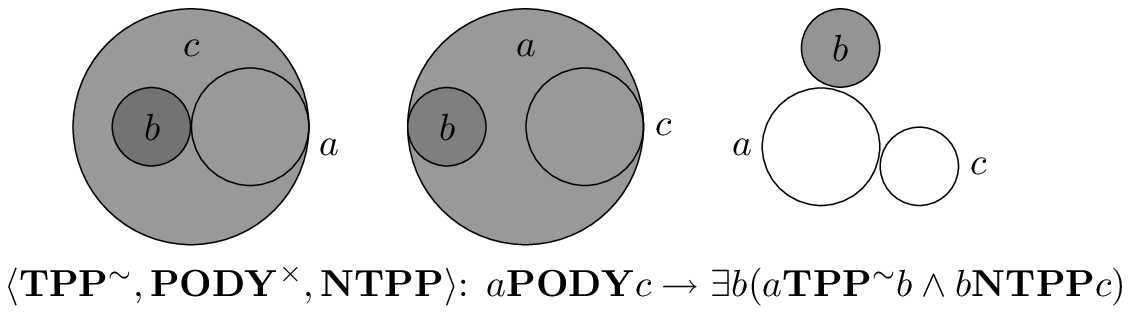}
\caption{\label{Fig:10x}Negative composition triads in Table
\ref{ReducedRCC11CT} are extensional on $D$.}
\end{figure}

\begin{prop}
 In the complemented closed disk algebra $\mathcal{L}$, the following composition
equations hold.

{\rm (C-1)\ \ $\TPP\circ \TPP=\TPP\cup \NTPP$;

(C-2)\ \  $\TPP\circ \TPPi=1^{\prime}\cup\TPP\cup\TPPi\cup
\PON\cup\ECN\cup\DC$;

(C-3)\ \ $\TPP\circ \NTPPi=\TPPi\cup
\NTPPi\cup\PON\cup\ECN\cup\DC$;

(C-4)\ \ $\TPP\circ \PON=\TPP\cup \NTPP\cup\PON\cup\ECN\cup\DC$;

(C-5)\ \ $\TPPi\circ \TPP=1^{\prime}\cup\TPP\cup
\TPPi\cup\PON\cup\PODY\cup\PODZ$;

(C-6)\  \ $\TPPi\circ \NTPP=\TPP\cup
\NTPP\cup\PON\cup\PODY\cup\PODZ$;

(C-7)\ \
$\TPPi\circ\PON=\TPPi\cup\NTPPi\cup\PON\cup\PODY\cup\PODZ$;

(C-8)\ \ $\NTPP\circ \NTPPi=1^{\prime}\cup\TPP\cup\TPPi\cup
\NTPP\cup\NTPPi\cup\PON\cup\ECN\cup\DC$;

 (C-9)\ \ $\NTPP\circ \PON=\TPP\cup \NTPP\cup\PON\cup\ECN\cup\DC$;

(C-10)\  $\NTPPi\circ
\NTPP=1^{\prime}\cup\TPP\cup\TPPi\cup\NTPP\cup\NTPPi
\cup\PON\cup\PODY\cup\PODZ$;

(C-11)\  $\NTPPi\circ \PON=\TPPi\cup
\NTPPi\cup\PON\cup\PODY\cup\PODZ$;

(C-12)\
$\PON\circ\PON=1^{\prime}\cup\TPP\cup\TPPi\cup\NTPP\cup\NTPPi
\cup\PON\cup\PODY\cup\PODZ\cup\ECN\cup\ECD\cup\DC$.}
\end{prop}
\begin{proof} Since regions in $D$ are either
 closed disks or the complement
of closed disks, the above equations can be verified using
elementary theory for circles (such as, internally tangent,
externally tangent, containment, disjoint, etc.). For each cell
entry in the reduced `extensional' RCC11 \textbf{CT} (Table
\ref{ReducedRCC11CT}) which is attached a superscript $^\times$,
we give an illustration for visual reference in Figure
\ref{Fig:10x}. Similar proofs can also be given to the other 61
triads.
\end{proof}

As a result, we know that the complemented closed disk algebra has
11 atoms and it's composition is just as the one given in the
RCC11 {\bf CT}.
\begin{thm}
The relation algebra determined by the RCC11 {\bf CT} can be
represented by the complemented closed disk algebra.
\end{thm}

\section{Summary and outlook}
This paper explored several important relation-algebraic questions
arising in the RCC theory. We have shown that the contact relation
algebra of $\mathfrak{B}_\omega$, a least RCC model, is not atomic
complete; and the contact relation algebra of the $n$-dimensional
Euclidean space is infinite and not integral. These results
suggest that in general we cannot obtain an extensional
composition table for the RCC theory by simply refining the RCC8
relations.

In order to obtain an extensional CT, one should restrict the
domain of regions: RCC models, in particular the regular closed
algebra of a regular connected space, might contain too much
regions. This has been partially demonstrated by the exhaustive
investigation of extensionality of RCC8 and RCC11 {\bf CT} given
respectively in \cite{LY03a} and Section 5 of the present paper.
There are also positive demonstrations. For the RCC8 table, the
closed disk algebra given in \cite{DUN} and the Egenhofer model
provides two extensional models which are arising from restriction
of regions in the real plane \cite{LY03b}. For the RCC11 table, we
have shown in Section 6 of this paper the complemented closed disk
algebra, whose domain contains only the closed disks and closures
of their complements in the real plane, is an extensional model.
Restricting the regions to connected regions bounded by Jordan
curves and closures of their complements seems also provide such a
model, this shall be investigated later.

Future work will investigate the contact relation algebra of
various small domains of regions which admits more operations than
complementation, e.g., finite unions or finite intersections. In
particular, the (complemented) Worboys-Bofakos model \cite{WB93}
deserves a detailed study with the tools of relation algebra. Note
that the 9-intersection principle can be applied to these domains,
we can compare the expressivity of RA logic with that of the
9-intersection model.


\end{document}